\documentclass[journal]{./IEEEtran}
\usepackage{amssymb}
\usepackage{amsmath}
\usepackage{amsfonts}
\usepackage{xcolor}
\usepackage{graphicx}
\usepackage{caption}
\usepackage{ragged2e}
\usepackage{cite}
\usepackage{booktabs}
\usepackage{tabularx}
\usepackage{makecell}
\usepackage{color}
\usepackage[linesnumbered,ruled,commentsnumbered]{algorithm2e}
\usepackage{amsmath}
\usepackage{multirow}
\usepackage{url}
\usepackage{tablefootnote}
\usepackage{subcaption}
\usepackage{hyperref}
\hypersetup{
colorlinks=true,
linkcolor=blue,
filecolor=purple,      
urlcolor=blue,
citecolor=green,
}

\begin{document}

\title{H2C: Hippocampal Circuit-inspired Continual Learning for Lifelong Trajectory Prediction in Autonomous Driving}

\author{Yunlong~Lin, Zirui~Li,~\IEEEmembership{Graduate Student Member,~IEEE},  Guodong Du, Xiaocong Zhao,~\IEEEmembership{Graduate Student Member,~IEEE},
        Cheng~Gong,~\IEEEmembership{Graduate Student Member,~IEEE}, Xinwei Wang,~\IEEEmembership{Member,~IEEE},
        Chao~Lu,~\IEEEmembership{Member,~IEEE},
        Jianwei~Gong,~\IEEEmembership{Memeber,~IEEE}
\thanks{This research is supported by the National Natural Science Foundation of China under Grant 52372405. (Yunlong Lin and Zirui Li contribute equally in this work.)}
\thanks{Yunlong Lin, Zirui Li, Guodong Du, Cheng Gong, Chao Lu and Jianwei Gong are with the School of Mechanical Engineering, Beijing Institute of Technology, Beijing 100081, China (Email: yunlonglin@bit.edu.cn; ziruili.work.bit@gmail.com; guodongdu\_robbie@163.com; chenggong@bit.edu.cn; chaolu@bit.edu.cn; gongjianwei@bit.edu.cn).}
\thanks{Xiaocong Zhao is with Key Laboratory of Road and Traffic Engineering, Ministry of Education, Tongji University, Shanghai, China (Email: zhaoxc@tongji.edu.cn).}
\thanks{Xinwei Wang is with the School of Engineering and Materials Science, Queen Mary University of London, UK (Email: xinwei.wang@qmul.ac.uk).}
 \thanks{Corresponding authors: Chao Lu and Jianwei Gong.}
}


\maketitle

\begin{abstract}

Deep learning (DL) has shown state-of-the-art performance in trajectory prediction, which is critical to safe navigation in autonomous driving (AD). However, most DL-based methods suffer from catastrophic forgetting, where adapting to a new distribution may cause significant performance degradation in previously learned ones. Such inability to retain learned knowledge limits their applicability in the real world, where AD systems need to operate across varying scenarios with dynamic distributions. As revealed by neuroscience, the hippocampal circuit plays a crucial role in memory replay, effectively reconstructing learned knowledge based on limited resources. Inspired by this, we propose a hippocampal circuit-inspired continual learning method (H2C) for trajectory prediction across varying scenarios. H2C retains prior knowledge by selectively recalling a small subset of learned samples. First, two complementary strategies are developed to select the subset to represent learned knowledge. Specifically, one strategy maximizes inter-sample diversity to represent the distinctive knowledge, and the other estimates the overall knowledge by equiprobable sampling. Then, H2C updates via a memory replay loss function calculated by these selected samples to retain knowledge while learning new data. Experiments based on various scenarios from the INTERACTION dataset are designed to evaluate H2C. Experimental results show that H2C reduces catastrophic forgetting of DL baselines by 22.71\% on average in a task-free manner, without relying on manually informed distributional shifts. The implementation is available at \href{https://github.com/BIT-Jack/H2C-lifelong}{https://github.com/BIT-Jack/H2C-lifelong}.

\end{abstract}


\IEEEpeerreviewmaketitle

\section{Introduction}\label{Section-I}

\begin{figure}[pt]

  \includegraphics[scale=1.0]{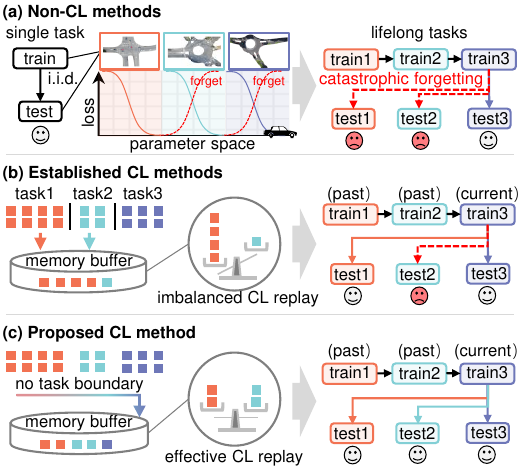}
  \caption{(a) DNN-based methods without continual learning (CL) suffer from catastrophic forgetting in lifelong trajectory prediction. (b) Most established CL methods either rely on informed task boundaries or exhibit an imbalanced memory replay. (c) The proposed CL method overcomes the imbalanced replay by integrating the diversity and randomness of the memory buffer, which is also independent of task boundary.}
  \label{fig1-intro}
\end{figure}

\IEEEPARstart{T}{rajectory} prediction supports safe and efficient planning of autonomous driving (AD) systems~\cite{pred-plan2025-its}. Most state-of-the-art (SOTA) trajectory prediction methods were developed based on deep neural networks (DNN) to forecast future intentions or positions of road users~\cite{diffusion2025-its}. However, as illustrated in Fig.~\ref{fig1-intro}(a), most DNN-based trajectory prediction methods are evaluated using testing samples that are independent and identically distributed (i.i.d.) to the training samples~\cite{mozaffari2020deep}. By contrast, AD systems may encounter varying scenarios with dynamic distributional shifts during their lifespan~\cite{kudithipudi2022biological-nature}. Once the DNN adapts to a new scenario, the prediction accuracy on previously learned scenarios may decline substantially~\cite{lesort2020cl-robot-survey}. This phenomenon occurs since the trainable parameters are optimized for new data, known as \emph{catastrophic forgetting}~\cite{forget}. In that case, inaccurate predictions may misguide the motion planning, harming the safety of AD~\cite{2025Haochen-TPAMI}.

To overcome the catastrophic forgetting without costly re-training or unlimited data maintenance, continual learning (CL)\footnote{Continual learning is also referred to as lifelong learning or incremental learning in much of the literature~\cite{chen2018lifelong, van2022three-incremental}.} has been proposed for DNN to learn dynamic distributions~\cite{TPAMI2024-comprehensive-cl}. In the paradigm of CL, training samples of different tasks with varying distributions arrive in a sequence. CL methods enable DNN to learn these sequential tasks with no or limited access to old samples and perform well in test sets of all encountered tasks~\cite{lesort2020cl-robot-survey}. Building upon this paradigm, many research fields have seen the advantage of CL methods to deal with ``lifelong tasks'' in artificial intelligence, including the incremental image classification~\cite{wang2023incorporating-nmi}, and lifelong place recognition~\cite{yin2023bioslamTRO}. In the scope of trajectory prediction, the lifelong tasks are defined as trajectory predictions across sequential scenarios~\cite{ma2021continual}, termed \emph{lifelong trajectory prediction}.

A few studies tried to mitigate the catastrophic forgetting in lifelong trajectory prediction using CL~\cite{ma2021continual, bao2023lifelong, lin2024continual}. Most of these CL methods for lifelong trajectory prediction are replay-based, which retain learned knowledge by approximating and recovering learned data distributions in past scenarios~\cite{TPAMI2024-comprehensive-cl}. Although they offer valuable insights, two main drawbacks limit their applicability in AD: First, as shown in Fig.~\ref{fig1-intro}(b), most established methods rely on manually informed task boundaries. However, the task boundary indicating the distributional shift is difficult for AD systems to detect in real-time since the environment is continuous. Second, most methods exhibit an imbalanced allocation of memory resources in replay-based strategies. With limited memory resources, the imbalance may lead to catastrophic forgetting of past tasks that have scarce data for replay. These problems raise the research question in this study: \emph{How to develop a lifelong trajectory prediction method that can effectively retain prior knowledge without accessing the information of task boundary?}

Recently, numerous biological fundamentals have been investigated to bridge the gap between neuroscience and artificial intelligence~\cite{zador2023catalyzing-neuroai}. Among CL-related biological underpinnings, the hippocampal circuit (HPC) inherently exhibits anti-forgetting characteristics~\cite{kudithipudi2022biological-nature}. Notably, recent findings further reveal that the HPC is crucial in processing complex continuous information~\cite{sun2025pattern}. These findings indicate the ``task-free''\footnote{Task-free is also termed task-agnostic in some studies~\cite{kudithipudi2022biological-nature, zeno2021task-agnostic, zhu2024tame}.} characteristic of HPC, which functions independent of the task boundary or task identifier (ID)\footnote{For the remainder of this paper, task boundary and task ID will be used interchangeably.}\cite{buzzega2020dark}. Specifically, the HPC efficiently represents continuous information by integrating two complementary mechanisms. First, to minimize overlapping representation of learned knowledge, similar information is distinguished by pattern separation~\cite{bakker2008pattern-science}. Second, pattern completion leverages sparse coding to construct a global representation for the continuous information~\cite{sun2025pattern}. Such integration incorporates the distinctive and overall representation in forming memory.

Drawing inspiration from these discoveries in neuroscience, this study proposes a \textbf{H}ippocampal \textbf{C}ircuit-inspired \textbf{C}ontinual learning (H2C) method by modeling the main functionalities of HPC to mitigate the catastrophic forgetting in lifelong trajectory prediction. As shown in Fig.~\ref{fig1-intro}(c), training samples from sequential scenarios arrive as a data stream. Without knowing the task boundary in the data stream, the H2C retains knowledge by selectively replaying a compact subset of learned samples. More specifically, the H2C employs two complementary strategies for the sample selection. As the distinctive representation, the first strategy emulates the pattern separation in HPC by maximizing inter-sample diversity. The second strategy mimics the pattern completion by performing uniform random sampling for the overall representation. Based on selected samples, the H2C replays such integrated representation to DNN via a task-free loss function to mitigate catastrophic forgetting. Compared with existing studies, the main contributions of this study are summarized as follows:

\begin{itemize}
\item[(1)] A novel CL methodology, H2C, is proposed to address the challenge of learning dynamic distributions in lifelong trajectory prediction. Inspired by HPC, the H2C selectively recovers learned knowledge via a task-free replay loss function, which mitigates the catastrophic forgetting of DNN-based trajectory predictors without accessing the task boundary.
\item[(2)] Two complementary sample selection strategies are developed for replay-based CL. The selected samples serve as memory resources to represent previously learned knowledge. By modeling the main functionalities of HPC, one strategy derives distinctive representation by maximizing inter-sample diversity. The other strategy employs an equiprobable sampling to capture the overall representation of learned distributions. The synergy between these two strategies effectively enhances the CL performance compromised by the imbalanced allocation of memory resources.
\item[(3)] Groups of experiments are designed to evaluate H2C. Based on various scenarios in a widely used benchmark of lifelong trajectory prediction, H2C is compared with five baselines in the experiments. Quantitative and qualitative experimental results with analysis are provided, demonstrating that the proposed H2C outperforms baselines in mitigating catastrophic forgetting and overall prediction accuracy for lifelong trajectory prediction.
\end{itemize}

The remainder of this article is organized as follows. Section \ref{Section-II} first introduces the related works to distinguish this study from previous works. Then, lifelong trajectory prediction is formulated in Section \ref{Section-III}. The proposed H2C is detailed in Section \ref{Section-IV}. Next, experimental settings, results, and discussion are presented in Section \ref{Section-V}, respectively. Finally, the conclusion and future works are summarized in Section \ref{Section-VI}.

\section{Related Works}\label{Section-II}
This study proposes a CL method for lifelong trajectory prediction in AD. We aim to mitigate the catastrophic forgetting of deep learning-based trajectory prediction. First, deep learning-based trajectory prediction methods with their advantages and shortcomings are reviewed. Then, we introduce the main theory of CL to overcome catastrophic forgetting by reviewing established CL methods. Finally, recent CL methods applied in AD are presented. By comparing these related works, research gaps are also summarized.

\subsection{Deep Learning-based Trajectory Prediction}\label{Section-II-A}

Deep learning-based trajectory prediction methods estimate future trajectories of road users (e.g., vehicles and pedestrians) by analyzing their motion states within an observation horizon, optionally augmented with map information~\cite{mozaffari2020deep, korbmacher2022deep-review}. Early methods were based on recurrent neural networks (RNN) and variants of RNN, including long short-term memory \cite{2016social-lstm, 2017RNN, Social-aware2022-pami} and gated recurrent unit~\cite{salzmann2020trajectron++}, which effectively model temporal dependencies in trajectory data. However, RNN struggled with multi-modal predictions, prompting the integration of generative methods~\cite{SEEM2023-pami}. For example, \cite{gupta2018socialGAN} generated diverse trajectory samples through adversarial training, achieving socially plausible multi-modal predictions. With the development of graph neural networks~\cite{li2021hierarchical} and attention mechanism~\cite{li2024UQnet}, spatial and social interactions were explicitly modeled by encoding relationships between road users as graphs~\cite{gu2021densetnt, graph2022-its}. Recent advancements leverage Transformer~\cite{vaswani2017Transformer} and diffusion models~\cite{bae2024singulartrajectory-diff} to parallelize long-range temporal and spatial modeling, achieving the SOTA performance for trajectory prediction~\cite{2024MTR++_PAMI, diffusion2025-its}.

These deep learning-based methods have demonstrated excellent performance under the i.i.d. assumption, where training and testing samples are independently drawn from the same distribution. However, most of them suffer from catastrophic forgetting in lifelong trajectory prediction. The inaccurate prediction in the forgotten scenarios may harm the safety of AD~\cite{pred-plan2025-its}.


\subsection{Continual Learning}\label{section-II-B}
CL, also referred to as lifelong learning~\cite{chen2018lifelong}, enables DNN to learn from a potentially infinite stream of data where all the data is not available at once~\cite{lesort2020cl-robot-survey}. From the aspect of the data distribution, CL can be characterized as learning from dynamic data distributions with limited access to observed training samples~\cite{TPAMI2024-comprehensive-cl}. The main challenge for the CL model is~\emph{catastrophic forgetting}, where adapting to a new task with the corresponding new distribution may lead to performance degradation in previously learned ones~\cite{forget}. Categorized by the mechanism, existing CL methods can be divided into three categories: Architecture-based, regularization-based, and replay-based CL. 

The architecture-based CL methods usually modify the structure of the models to construct task-specific parameters to retain learned knowledge~\cite{wang2017growing, mallya2018packnet, mallya2018piggyback}. The regularization-based CL methods protect model performance in old tasks by adding regularization terms to constrain parameter updates~\cite{kirkpatrick2017EWC, li2017learningLwF}. However, as the number of tasks grows, the model complexity of DNN will increase significantly in architecture-based methods, and DNN may saturate due to excessive regularization in regularization-based methods~\cite{lesort2020cl-robot-survey}. By contrast, replay-based CL mitigates catastrophic forgetting by approximating and recovering previously learned data distributions~\cite{TPAMI2024-comprehensive-cl}. \cite{lesort2019generative-dgr} generated pseudo samples to approximate the observed samples from old tasks. Differently, \cite{lopez2017gradient-gem} and \cite{chaudhry2019efficient-agem} proposed to partially store observed samples from old tasks. Based on the stored samples, DNN models were trained under an inequality constraint, which mitigated catastrophic forgetting by restricting the increment of losses in old tasks. Furthermore, \cite{buzzega2020dark} partially stored observed samples with output logits using reservoir sampling~\cite{vitter1985random-rsvr}, mitigating catastrophic forgetting by encouraging DNN to mimic the output logits in old tasks. These replay-based CL methods offer insights in efficiently mitigating catastrophic forgetting. However, most of them exhibit imbalanced CL replay, where old tasks are represented by imbalanced memory resources such as stored or generated samples. Since the memory resources are limited, such imbalance will lead to performance degradation in CL for some old tasks with a small amount of replayed data.

\subsection{Continual Learning in Autonomous Driving}\label{Section-II-C}

According to \cite{sae2014taxonomy}, fully AD is required to be capable of operating across all driving conditions and scenarios. Towards such requirements for real-world application, learning-based AD systems need to incrementally learn and remember knowledge in different tasks during their lifespan~\cite{lesort2020cl-robot-survey}. Recent efforts based on CL have emerged to enhance the applicability of AD. \cite{2025-humanguided-CL} employed the regularization-based CL method~\cite{kirkpatrick2017EWC} for AD decision-making. Similarly, a regularization-based CL method was designed for lifelong pedestrian trajectory prediction in \cite{2022-EWC-ped-traj}. For replay-based CL methods, \cite{GC2024-lifelong} addressed lifelong path tracking in changing environments by using the method~\cite{chaudhry2019efficient-agem}. By revising the method~\cite{lopez2017gradient-gem}, \cite{lin2024continual} improved the computational efficiency of CL for lifelong vehicle trajectory prediction. Inspired by the replay-based CL method~\cite{shin2017continual-DGR}, \cite{ma2021continual} and \cite{bao2023lifelong} mitigated catastrophic forgetting in vehicle trajectory prediction by generating pseudo data to approximate learned knowledge in old tasks. These CL-based explorations took steps forward to the high-level AD~\cite{li2023continual-survey}. However, most of these CL methods applied in AD still rely on the task boundary to identify the distributional shift between tasks~\cite{lin2023rethinking}. For example, the allocation of memory resources depends on knowing every task shift in~\cite{lin2024continual}. The generation of pseudo samples also relies on the task boundary in~\cite{ma2021continual}. However, it is challenging to identify the potential data distribution behind each task from the naturalistic driving data collected in the real world. Therefore, CL methods for trajectory prediction are expected to be independent of task boundary, termed task-free~\cite{he2024dyson} or task-agnostic~\cite{kudithipudi2022biological-nature}. 

To address the limitations of the abovementioned studies, the proposed H2C allows deep learning-based trajectory prediction methods to retain accurate prediction across distributional shifts without re-training or requiring the vast amount of data available simultaneously. The core algorithms are two complementary strategies for sample selection, overcoming the imbalanced CL replay. Besides, the CL mechanism within H2C is task-free, which do not rely on task boundary information during training.

\begin{figure*}[htbp]
      \centering
      \includegraphics[scale=1.0]{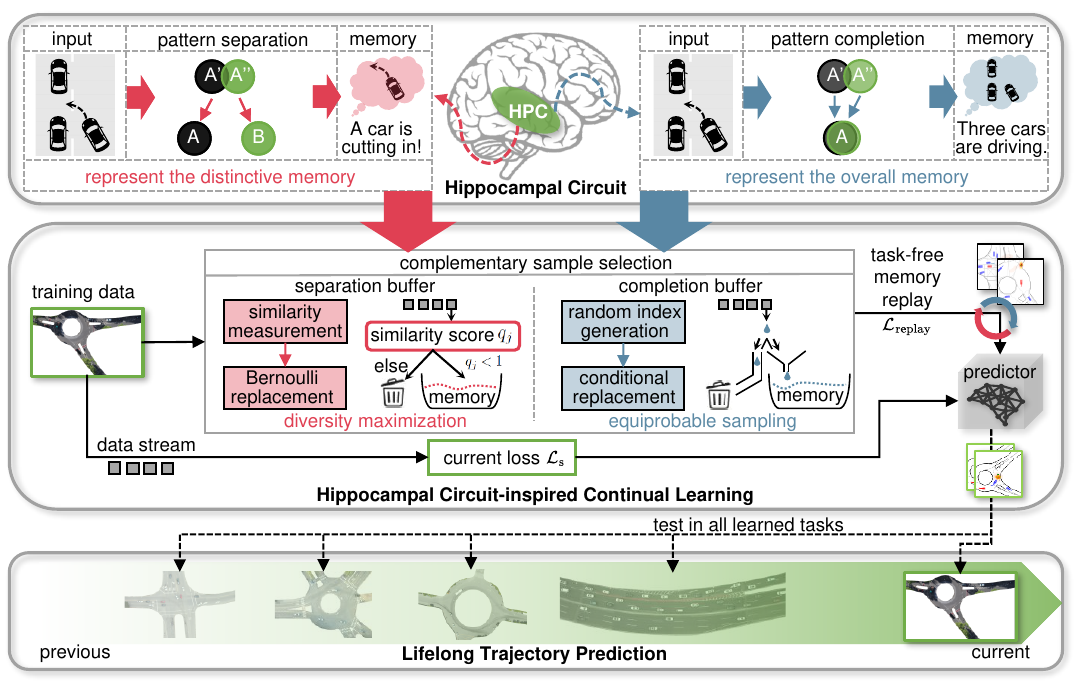}
      \caption{Schematic of the proposed H2C for lifelong trajectory prediction.}
      \label{fig_methodology}
\end{figure*}

\section{Problem Formulation}\label{Section-III}

Lifelong trajectory prediction models the trajectory prediction in dynamic distributions across different scenarios. In this section, we will first formulate the learning-based trajectory prediction as the preliminary. Then, the problem formulation of the lifelong trajectory prediction is presented.

\subsection{Learning-based Trajectory Prediction}\label{section-III-A}

Trajectory prediction for AD aims to estimate future positions or intentions of road users (e.g., vehicles, pedestrians, and cyclists). Specifically, the predicted road users are defined as target agents, while the others, that can influence the behavior of the target agents, are termed surrounding agents. Learning-based trajectory prediction models leverage historical trajectories of both the target agent and its surrounding agents to achieve accurate forecasting.


Formally, if a DNN-based prediction model parameterized by $\boldsymbol{\theta}$ is denoted as $f_{\boldsymbol{\theta}}$, the learning-based trajectory prediction under a distribution $\mathcal{D}$ can be formulated as:
\begin{equation}
    \hat{\mathbf{Y}}^{\mathcal{D}} = f_{\boldsymbol{\theta}}\left ( \mathbf{X}^{\mathcal{D}} \right ) \label{eq_output}
\end{equation}
In \eqref{eq_output}, the input at the current time step $t_{\text{c}}$ over an observation horizon $t_{\text{obs}}$ is presented as $\mathbf{X}^{\mathcal{D}}=[\mathbf{X}_{t_{\text{c}}-t_{\text{obs}}+1}, ..., \mathbf{X}_{t_{\text{c}}}]$, where $\textbf{X}_{t}$ contains state features (e.g., positions and velocities) of the target and surrounding agents, and map features (e.g., road boundary and centerline of lanes) at time $t$. Given a prediction horizon $t_\text{pred}$, the output $\hat{\mathbf{Y}}^{\mathcal{D}}$ estimates the future positions $\mathbf{Y}^{\mathcal{D}}$ of the target agent at time $t_\text{c}+t_\text{pred}$.

\subsection{Lifelong Trajectory Prediction}\label{Section-III-B}
The lifelong trajectory prediction is characterized as a sequence of learning-based trajectory prediction tasks $\{\text{T}_1, ..., \text{T}_{i}, ..., \text{T}_N\}$, where $N \in \mathbb{Z}^{+}$ denotes the total number of tasks. Following~\cite{ma2021continual, bao2023lifelong, lin2024continual}, the $i$\textsuperscript{th} task $\text{T}_{i}$ is defined as the learning-based trajectory prediction using the dataset collected in the $i$\textsuperscript{th} scenario under a distribution $\mathcal{D}_{\text{T}_i}$. The distribution $\mathcal{D}_{\text{T}_i}$ varies between different tasks, and the DNN-based model $f_{\boldsymbol{\theta}}$ aims to learn dynamic distributions $\{\mathcal{D}_{\text{T}_1}, ...,\mathcal{D}_{\text{T}_i}, ..., \mathcal{D}_{\text{T}_N}\}$ sequentially.

Using the subscript $j$ to index samples, the data stream of the lifelong trajectory prediction can be represented as:
\begin{equation}
    \mathbf{S} = \left \{ (\mathbf{X}_j^{\mathcal{D}_{\text{T}_{i}}}, \mathbf{Y}_j^{\mathcal{D}_{\text{T}_i}}) | 1 \le i \le  N, 1 \le j \le N_{\text{T}_i}  \right \}  \label{eq_datastream}
\end{equation}
where $N_{\text{T}_i}$ is the number of i.i.d. samples drawn from the distribution $\mathcal{D}_{\text{T}_i}$. Since the model encounters $N$ tasks $\{\text{T}_1, ..., \text{T}_{i}, ..., \text{T}_N\}$ sequentially, the training samples from all tasks are not accessible simultaneously. Ordered by the task, training samples are gradually available from \eqref{eq_datastream}. Meanwhile, the data stream $\mathbf{S}$ is for one-time observation, where the sample is abandoned once it is observed by the model. Notably, if a CL method needs to re-use observed samples based on a memory buffer $\mathcal{M}$ (e.g., replaying the stored samples), the buffer size $|\mathcal{M}|$, i.e., the maximum amount of samples for re-usage, is also limited. Referring to~\cite{lesort2020cl-robot-survey}, the relationship between the amount of total samples in \eqref{eq_datastream} and $|\mathcal{M}|$ follows the constraint:
\begin{equation}
   | \mathcal{M} | \ll \sum_{i=1}^{N}N_{\text{T}_i} \label{eq_amount}
\end{equation}

Under these assumptions, the learning objective of the lifelong trajectory prediction can be formulated as:
\begin{equation}
    \boldsymbol{\theta}^{*} = \arg\min_{\boldsymbol{\theta}} \sum_{i=1}^{N}\sum_{j=1}^{N_{\text{T}_{i}}} \ell\left( f_{\boldsymbol{\theta}}\left( \mathbf{X}_{j}^{\mathcal{D}_{\text{T}_i}}, \mathcal{M} \right), \mathbf{Y}_{j}^{\mathcal{D}_{\text{T}_i}} \right)
    \label{eq_objective}
\end{equation}
where $\ell (\cdot )$ is the loss function. The main challenge to achieve the objective \eqref{eq_objective} lies in the assumption that observed samples cannot be fully accessed simultaneously. Moreover, the task ID $\{\text{T}_1, ..., \text{T}_{i}, ..., \text{T}_N\}$ is assumed to be unavailable for task-free CL methods during training.

\section{Hippocampal Circuit-inspired Continual Learning}\label{Section-IV}

The overview of the proposed H2C is demonstrated in Fig.~\ref{fig_methodology}. Inspired by the pattern separation and pattern completion in HPC, the complementary sample selection strategies partially store observed training samples to represent learned knowledge. Based on the stored samples, H2C enables the DNN-based predictor to retain learned knowledge via the task-free memory replay while continuously updating with new data. In this section, we will briefly introduce the biological inspiration of the proposed method. Then, the complementary selection strategies and the task-free memory replay are detailed.

\subsection{Pattern Separation and Pattern Completion in Hippocampal Circuit}\label{Section-IV-A}

\begin{figure}[bp]
      \centering
      \includegraphics[scale=1.0]{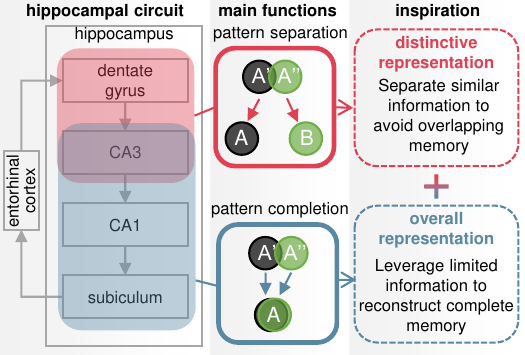}
      \caption{The inspiration from pattern separation and pattern completion in the HPC. The schematic of HPC is modified from the paper \cite{sun2025pattern}.}
      \label{fig_HPC}
\end{figure}

HPC in the human brain plays a pivotal role in forming memory, showing advantages in processing continuous information~\cite{sun2025pattern}. As depicted in Fig.~\ref{fig_HPC}, pattern separation and pattern completion are two main functionalities of HPC. Pattern separation, occurring primarily in the dentate gyrus and CA3 subfield, distinguishes similar information by generating distinctive neural representations. This prevents the memory interference caused by representational overlapping. Supported by CA3, CA1, and the subiculum, pattern completion can rebuild complete memory based on limited information.

As two complementary mechanisms in HPC, pattern separation reflects the distinctiveness of representations, highlighting the diversity in the memory. From the aspect of overall representation, pattern completion uses limited information to represent the complete memory. Such collaborative mechanisms enable the HPC to maintain the overall integrity of memory representations while ensuring their distinctiveness.

Inspired by these mechanisms in HPC, two complementary sample selection strategies are developed to represent learned knowledge in lifelong trajectory prediction. Building upon the first strategy that models the functionality of pattern separation, we construct the separation buffer to represent distinctive learned knowledge. Similar to pattern completion, the completion buffer is constructed based on the other strategy to capture the overall representation of learned knowledge in lifelong tasks. More details are presented in Section~\ref{Section-IV-B}.

\subsection{Complementary Sample Selection}\label{Section-IV-B}
 
In replay-based CL, memory buffers are responsible to selectively store a small subset of observed samples, which are used to represent learned knowledge. Modeling the functionalities of pattern separation and pattern completion in HPC, two complementary sample selection strategies are developed to construct two memory buffers.

\subsubsection{Separation Buffer}
For the distinctive representation of learned knowledge, the separation buffer is constructed by maximizing the diversity of samples. As distinguishing similar information in pattern separation, the separation buffer first measures the diversity of encountered samples. In detail, the buffer tends to select distinctive samples by comparing the diversity between the newly encountered sample and the stored ones. The diversity is measured based on the cosine similarity between loss gradients of samples~\cite{aljundi2019-gss}, termed as similarity score $q$. Let $\boldsymbol{g}$ denote the loss gradient of the newly observed sample. We randomly compare loss gradients of $B$ memory samples within the separation buffer $\mathcal{M}_\text{sp}$, the corresponding similarity score $q$ is calculated by:
\begin{equation}
    q=\max_{b\in\{1,...,B\}}\left ( \frac{\left \langle \boldsymbol{g},\boldsymbol{g}_{b}  \right \rangle }{\left \| \boldsymbol{g}  \right \| \left \| \boldsymbol{g}_b \right \|  }  \right )+1 \label{eq_score}
\end{equation}
where $\boldsymbol{g}_b$ is the loss gradient of the $b$\textsuperscript{th} memory samples for comparison. $\left \langle \cdot \right \rangle $ and $\left \| \cdot \right \| $ denote the inner product and the modulus of the gradients, respectively. The sample stored in the buffer with the larger $q$ is more likely replaced by the newly encountered one. With a limited buffer size $|\mathcal{M_\text{sp}}|$, the separation buffer aims to maximize the diversity of stored samples.

\SetAlgoNlRelativeSize{0} %
\SetNlSty{}{}{} %
\SetAlgoNlRelativeSize{-1} %
\SetCommentSty{textnormal} %
\SetKwComment{tcp}{$\triangleright$}{}

\begin{algorithm}
\caption{Selection Strategy for Pattern Separation}\label{al_completion}

\KwIn{The data stream $\mathbf{S}$, where the $j\textsuperscript{th}$ sample is denoted as $\boldsymbol{d}_j$ with the similarity score $q_j$; the separation buffer $\mathcal{M}_\text{sp}$ with the buffer size $|\mathcal{M}_\text{sp}|$; the number of samples $B$ for score computation; the loss function $\ell$; the trajectory prediction model ${f_{\boldsymbol{\theta}}}$.}
\KwOut{The updated $\mathcal{M}_\text{sp}$, where the $i\textsuperscript{th}$ stored sample is denoted as $\mathcal{M}_\text{sp}(i)$.}

\For{$j$ \textbf{in} range(1, $|\mathbf{S}|$)}{
    \eIf{$j == 1$}{
        $q_j \gets 0.1$\tcp*[r]{Initialization.}
    }{
        $\boldsymbol{g} \gets \nabla \ell(f_{\boldsymbol{\theta}}(\boldsymbol{d}_j))$\tcp*[r]{Calculate gradients.}
        $\Omega \gets \emptyset$\;
        
        \For{$i$ \textbf{in} range(1, $B$)}{
            $b \sim \text{Uniform}(1, |\mathcal{M}_\text{sp}|)$\;
            $\Omega \gets \Omega \cup \mathcal{M}_\text{sp}(b)$\tcp*[r]{Randomly select $B$ stored samples for the score computation.}
        }
        
        \For{all samples in $\Omega$}{
            $\boldsymbol{g}_b \gets \nabla \ell(f_{\boldsymbol{\theta}}(\mathcal{M}_\text{sp}(b)))$\;
        }
        
        $q_j \gets \max_b \left( \frac{\langle \boldsymbol{g}, \boldsymbol{g}_b \rangle}{\|\boldsymbol{g}\|\|\boldsymbol{g}_b\|} \right) + 1$\tcp*[r]{Similarity score.}
        
        \eIf{$j \leq |\mathcal{M}_\mathrm{sp}|$}{
            $\mathcal{M}_\text{sp}(j) \gets \boldsymbol{d}_j$\tcp*[r]{Add sample before full.}
            $\mathcal{M}_\text{sp} \gets \mathcal{M}_\text{sp} \cup \mathcal{M}_\text{sp}(j)$\tcp*[r]{Update.}
        }
        {\If{$q_j < 1$}{
            $i \sim \mathbb{P}(i) = \frac{q_i}{\sum_{i=1}^{|\mathcal{M}_\text{sp}|} q_i}$\;
            $r \sim \text{Uniform}(0, 1)$\;
            \If{$r < \frac{q_i}{q_i + q_j}$}{
                $\mathcal{M}_\text{sp}(i) \gets \boldsymbol{d}_j$\tcp*[r]{Update.}
                $q_i \gets q_j$\;}
            }
        }
    }
}
\end{algorithm}

As shown in Algorithm~\ref{al_completion}, samples from the data stream formulated in \eqref{eq_datastream} are sequentially input to the method. When the separation buffer $\mathcal{M}_\text{sp}$ is not full, each newly encountered sample $\boldsymbol{d}_j$ is directly stored with its similarity score $q_j$. Once the buffer is full, a sample will be randomly selected from the buffer as the candidate to be replaced. Let $q_i$ denote the similarity score of the candidate. In that case, the replacement is a Bernoulli event with probability $P_\text{replace}$ to happen:
\begin{equation}
    P_\text{replace}=\frac{q_i}{q_i+q_j} 
\end{equation}

\subsubsection{Completion Buffer}
Similar to pattern completion which uses partial information to reconstruct the complete memory, the completion buffer builds an overall representation of learned distributions via equiprobable sampling. Specifically, the completion buffer aims to randomly select samples, where each sample from the data stream has an equal probability of being stored in the buffer. However, the total length of the data stream $|\mathbf{S}|$ is unknown since the sample is available incrementally. In this situation, widely used strategies for equiprobable sampling such as Fisher-Yates shuffle are inapplicable since they require the prior information of $|\mathbf{S}|$. As an efficient solution, we apply the reservoir sampling to obtain an unbiased estimation of learned distributions~\cite{vitter1985random-rsvr}. 

As depicted in Algorithm~\ref{al_1}, the completion buffer is denoted as $\mathcal{M}_\text{cp}$ with the buffer size $|\mathcal{M}_\text{cp}| \in \mathbb{Z}^{+}$. First, the completion buffer is initialized by an empty set. Then, every observed sample is stored in the buffer before it is full. After that, the newly encountered sample may replace one of the stored samples in the buffer. Note that the probability to be stored is expected to be equal to all samples in the data stream. In detail, $\forall j > |\mathcal{M}_\text{cp}|$, the $j\textsuperscript{th}$ observed sample is selected as the candidate with the probability of $|\mathcal{M}_\text{cp}|/j$. Once the $j\textsuperscript{th}$ sample becomes the candidate, one of the previously stored samples will be replaced by the candidate with the probability of $1/|\mathcal{M}_\text{cp}|$. Let $\mathcal{M}_\text{cp}(i)$ denote the $i$\textsuperscript{th} sample stored in the completion buffer. The update step is implemented via a conditional replacement:
\begin{equation}
    \mathcal{M}_\text{cp}(r) = \begin{cases} 
\boldsymbol{d}_j, & \text{if } r \le |\mathcal{M}_\text{cp}|,  r \sim \mathrm{Uniform}(1,j) \\
\mathcal{M}_\text{cp}(r), & \text{otherwise.}
\end{cases}
\end{equation}
where $r\in\mathbb{Z}^{+}$ is the randomly generated index. As a result, the probability to be stored in the buffer is $|\mathcal{M}_\text{cp}|/|\mathbf{S}|$ for each sample in $\mathbf{S}$.

In summary, the separation buffer constructs a distinctive representation of learned knowledge by maximizing the diversity of samples. The overall representation is achieved by the completion buffer via an equiprobable sampling. Besides, these sample selection strategies is designed for the incrementally available data stream, which do not need the information of task ID. Based the two memory buffers, we recall the distinctive and overall memory to the DNN via a replay mechanism, which is detailed in Section~\ref{Section-IV-C}.

\begin{algorithm}[tp]
\SetAlgoNlRelativeSize{0}  
\SetNlSty{textbf}{}{} 
\SetAlgoNlRelativeSize{-1} 
\SetCommentSty{textnormal} 
\SetKwComment{tcp}{$\triangleright$}{}

\caption{Selection Strategy for Pattern Completion}\label{al_1}

\KwIn{The data stream $\mathbf{S}$, where the $j\textsuperscript{th}$ sample is denoted as $\boldsymbol{d}_j$; the completion buffer $\mathcal{M}_\text{cp}$ with the buffer size $|\mathcal{M}_\text{cp}|$.}
\KwOut{The updated $\mathcal{M}_\text{cp}$, where the $i\textsuperscript{th}$ sample stored in $\mathcal{M}_\text{cp}$ is denoted as $\mathcal{M}_\text{cp} (i)$.}

$\mathcal{M}_\text{cp} \gets \emptyset$\tcp*[r]{Initialization.}

\For{$j$ \textbf{in} range(1, $|\mathbf{S}|$)}{
    \eIf{$j \leq |\mathcal{M}_\mathrm{cp}|$}{
        $\mathcal{M}_\text{cp}(j) \gets \boldsymbol{d}_j$\tcp*[r]{Add sample before full.}
        $\mathcal{M}_\text{cp} \gets \mathcal{M}_\text{cp} \cup \mathcal{M}_\text{cp}(j)$\tcp*[r]{Update.}
    }{
        $r \sim \text{Uniform}(1, j)$\;
        \If{$r \leq |\mathcal{M}_\mathrm{cp}|$}{
            $\mathcal{M}_\text{cp}(r) \gets \boldsymbol{d}_j$\tcp*[r]{Update.}
        }
    }
}
\end{algorithm}

\subsection{Task-free Memory Replay}\label{Section-IV-C}
Towards the sequentially available data stream $\mathbf{S}$ in \eqref{eq_datastream}, the H2C aims to achieve the objective formulated in \eqref{eq_objective} without knowing the task ID. For convenience, the encountered sample without the task ID is denoted as $(\mathbf{X}, \mathbf{Y})$. At each step of model optimization, the training loss on the newly encountered sample from the data stream $\mathbf{S}$ is represented as:
\begin{equation}
    \mathcal{L}_\text{s} = \ell \left (  f_{\boldsymbol{\theta}} \left ( \mathbf{X} \right ),\mathbf{Y}  \right ) \label{eq_loss_current}
\end{equation}

As described in Section~\ref{Section-III-B}, the encountered sample is either abandoned or stored in the memory buffer after being observed. The sample stored in the memory buffers is denoted as $(\mathbf{X}^{'}, \mathbf{Y}^{'})$, termed memory sample. Moreover, let $f_{\boldsymbol{\theta_{\text{init}}}}$ denote the state of the DNN-based model when it observes the sample $(\mathbf{X}^{'},\mathbf{Y}^{'})$. We also store the initial output $\hat{\mathbf{Y}}^{'}_{\text{init}}=f_{\boldsymbol{\theta_\text{init}}}(\mathbf{X}^{'})$ of each memory sample. The triplets $(\mathbf{X}^{'},\mathbf{Y}^{'},\hat{\mathbf{Y}}^{'}_{\text{init}})$ stored in the memory buffer are used for the task-free memory replay, where the loss is calculated based on the memory samples. Meanwhile, the model is also encouraged to mimic its initial response to the memory sample, i.e., $\hat{\mathbf{Y}}^{'}_{\text{init}}$. Let $\mathcal{M}$ denote the memory buffer. The proposed task-free memory replay can be formulated as optimizing the model using the following loss function:
\begin{equation}
\begin{aligned}
    \mathcal{L}_\text{replay}(\mathcal{M}) &= \mathbb{E}_{(\mathbf{X}^{'},\mathbf{Y}^{'})\sim \mathcal{M}}
    \left[ \ell(f_{\boldsymbol{\theta}}(\mathbf{X}^{'}),\mathbf{Y}^{'}) \right] \\
    &\quad + \mathbb{E}_{(\mathbf{X}^{'},\hat{\mathbf{Y}}^{'}_\text{init})\sim \mathcal{M}}
    \left[ D_\text{KL}\left(f_{\boldsymbol{\theta}} (\mathbf{X}^{'}) \left| \right| \hat{\mathbf{Y}}^{'}_{\text{init}} \right) \right]
\end{aligned}\label{eq_replay_origin}
\end{equation}
where $\mathbb{E}$ is the mathematical expectation, and $D_\text{KL} (\cdot)$ denotes the Kullback-Leibler divergence. As suggested by~\cite{2015Distilling}, the optimization of the Kullback-Leibler divergence in \eqref{eq_replay_origin} can be simplified equivalently as minimizing the squared Euclidean distance under mild assumptions. For a more efficient calculation, \eqref{eq_replay_origin} is transformed as follows:
\begin{equation}
\begin{aligned}
    \mathcal{L}_\text{replay}(\mathcal{M}) &= \mathbb{E}_{(\mathbf{X}^{'},\mathbf{Y}^{'})\sim \mathcal{M}}
    \left[ \ell(f_{\boldsymbol{\theta}}(\mathbf{X}^{'}),\mathbf{Y}^{'}) \right] \\
    &\quad + \mathbb{E}_{(\mathbf{X}^{'},\hat{\mathbf{Y}}^{'}_{\text{init}})\sim \mathcal{M}}
    \left[ \| f_{\boldsymbol{\theta}} (\mathbf{X}^{'}) - \hat{\mathbf{Y}}^{'}_\text{init} \|_2^2 \right]
\end{aligned}\label{eq_replay_rewritten}
\end{equation}

Note that the memory buffer is denoted as $\mathcal{M}$ in \eqref{eq_replay_rewritten}, which is replaced by the separation buffer $\mathcal{M}_\text{cp}$ or the completion buffer $\mathcal{M}_\text{sp}$ in the implementation. 
Finally, the DNN-based model $f_{\boldsymbol{\theta}}$ is trained by the loss on the encountered sample with the task-free memory replay. The total loss is represented as:
\begin{equation}
    \mathcal{L}_\text{total}=\mathcal{L}_\text{s}+\alpha \mathcal{L}_\text{replay}(\mathcal{M}_\text{sp})+\beta \mathcal{L}_\text{replay}(\mathcal{M}_\text{cp}) \label{eq_total_loss}
\end{equation}
where $\alpha$ and $\beta$ are coefficients for losses based on the separation buffer and completion buffer, respectively.

\section{Experiments}\label{Section-V}
The primary objective of this study is to mitigate catastrophic forgetting in lifelong trajectory prediction. The proposed H2C retains learned knowledge using the complementary sample selection with task-free memory replay. 
Experiments based on the INTERACTION dataset are designed to evaluate H2C~\cite{zhan2019interaction-dataset}. We use diverse metrics to measure prediction accuracy and anti-forgetting performance in the evaluation. To further validate the advantages of H2C, we compare H2C against five established methods in the experiments, providing quantitative and qualitative analysis.

\subsection{Datasets and Basic Model}\label{Section-V-A}
INTERACTION dataset contains diverse sub-datasets collected from different scenarios~\cite{zhan2019interaction-dataset}, which is widely adopted as the benchmark for lifelong trajectory prediction. We construct four groups of sequential scenarios based on the INTERACTION dataset. Information on the used sub-datasets and the corresponding bird's-eye-view photos are demonstrated in Table~\ref{table_datasets} and Fig.~\ref{fig-datasets}, respectively.

\begin{figure}[tp]
  \centering

  \includegraphics[scale=1.0]{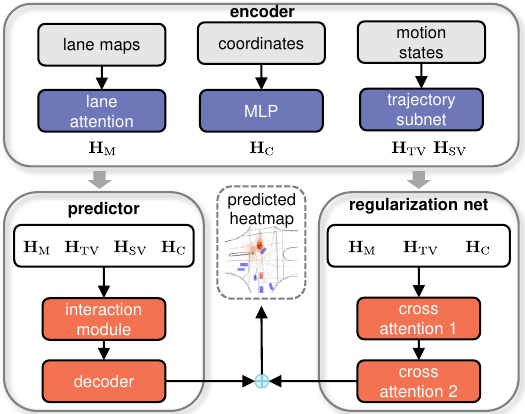}
  \caption{The basic model implemented in experiments.}
  \label{fig3_model}
\end{figure}

Taking vehicles as an example of road users, vehicle trajectory prediction is implemented in the experiments. Corresponding to the target agent and surrounding agents defined in Section~\ref{Section-III}, vehicles in the dataset are divided into the target vehicle (TV) and surrounding vehicles (SVs).

The basic model implemented in experiments is the uncertainty quantification network (UQnet)~\cite{li2024UQnet}, which is the SOTA for vehicle trajectory prediction in the open challenge\footnote{INTERPRET: Interaction-dataset-based prediction challenge, single-agent track, organized by ICCV 2021 Competition. Available at: \url{https://challenge.interaction-dataset.com/leader-board/}. Last accessed on April 2\textsuperscript{nd}, 2025.}. The H2C and all compared methods are applied to the UQnet in the experiment. The structure of the UQnet is shown in Fig.~\ref{fig3_model}. In detail, the attention-based module encodes the connectivity feature of roads, denoted as $\mathbf{H}_\text{M}$. The coordinates of the mesh grid are encoded by the multi-layer perceptron (MLP), denoted as $\mathbf{H}_\text{C}$. Besides, $\mathbf{H}_\text{S}$ and $\mathbf{H}_\text{TV}$ represent the encoded motion states of SVs and TV, respectively. Then, the extracted features are passed into the regularization net and the DenseTNT-based predictor~\cite{gu2021densetnt}. Based on the regularization term of the focal loss~\cite{2020focal_loss} calculated by the regularization net, the UQnet predicts the future position of the TV. 

Specifically, the prediction is a two-dimensional (2D) spatial distribution described by a mesh-grid heatmap $\hat{\bf{Y}}^{\text{hm}}$. Assuming that the 2D heatmap has $h$ rows and $w$ columns and indexing the position of grids in the heatmap by subscripts $i \in [1, h]$ and $j \in [1, w]$, each value $\hat{\bf{Y}}_{i,j}^{\text{hm}}$ in $\hat{\bf{Y}}^{\text{hm}}$ is the estimated probability that the TV will be present within that specific grid. Furthermore, $W$ predicted endpoints of TV are derived from the heatmap using a naive local-maximum sampling strategy~\cite{gilles2022gohome}. In the experiments, we set $W=6$. For more details of the UQnet, please refer to \cite{li2024UQnet}.


\begin{figure*}[ht]
  \centering
  \includegraphics[scale=1.0]{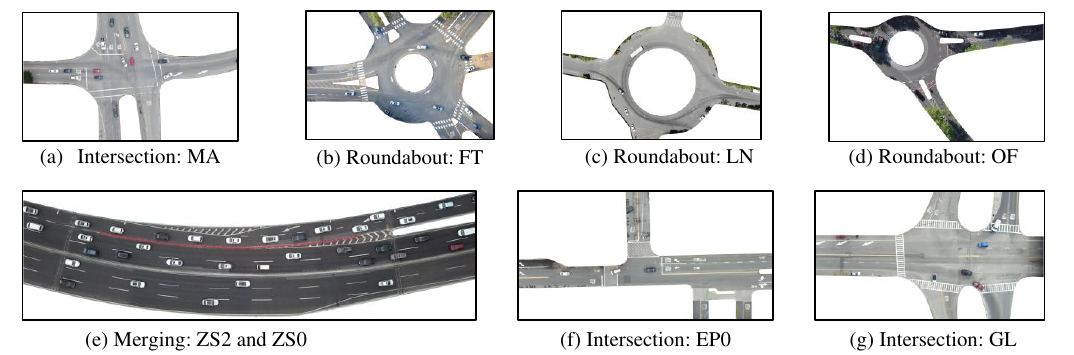}
  \caption{\justifying{Real-world scenarios used in experiments from a bird's-eye-view~\cite{zhan2019interaction-dataset}. (a) A non-signalized intersection in the U.S.A. is denoted as MA. (b) A busy 7-way roundabout in the U.S.A. is denoted as FT. (c) A roundabout in China is denoted as LN, where all branches are controlled by ``yield" signs. (d) A roundabout in Germany is denoted as OF, where all branches are controlled by ``yield" signs. (e) A highway merging scenario in China. ZS0 refers to the upper lanes with a zipper merging, and ZS2 refers to the lower lanes with a lane-change merging. (f) A busy all-way-stop T-intersection in the U.S.A. is denoted as EP0. (g) A non-signalized intersection in the U.S.A. is denoted as GL.}}
  \label{fig-datasets}
\end{figure*}

\begin{table}[tp]
    \centering
    \caption{The Sub-datasets in INTERACTION dataset}
    \label{table_datasets}
    \begin{tabularx}{\linewidth}{>{\centering\arraybackslash}p{0.07\linewidth} >{\centering\arraybackslash}p{0.06\linewidth} >{\centering\arraybackslash}p{0.14\linewidth} >{\centering\arraybackslash}p{0.11\linewidth} | *{4}{>{\centering\arraybackslash}X}}
    \toprule
    Task ID & Sub-dataset & Scenario type & Training samples & Group  \uppercase\expandafter{\romannumeral1} & Group  \uppercase\expandafter{\romannumeral2}  & Group  \uppercase\expandafter{\romannumeral3}  & Group  \uppercase\expandafter{\romannumeral4}   \\ 
    \midrule
     $\text{T}_1$ & MA & Intersection & 33,456  & $\checkmark$ & $\checkmark$ & $\checkmark$ & $\checkmark$ \\
     $\text{T}_2$ & FT & Roundabout & 66,256    & $\checkmark$ & $\checkmark$ & $\checkmark$ & $\checkmark$ \\
     $\text{T}_3$ & LN & Roundabout & 3,400    & $\checkmark$ & $\checkmark$ & $\checkmark$ & $\checkmark$ \\
     $\text{T}_4$ & ZS2 & Merging & 15,400      & $\checkmark$ & $\checkmark$ & $\checkmark$ & $\checkmark$ \\
     $\text{T}_5$ & OF & Roundabout & 7,904    & $\checkmark$ & $\checkmark$ & $\checkmark$ & $\checkmark$ \\
     $\text{T}_6$ & EP0 & Intersection & 9,136 & $\times$ & $\checkmark$ & $\checkmark$ & $\checkmark$ \\
     $\text{T}_7$ & GL & Intersection & 81,640  & $\times$ & $\times$  & $\checkmark$ & $\checkmark$ \\
     $\text{T}_8$ & ZS0 & Merging & 35,512      & $\times$ & $\times$ & $\times$ & $\checkmark$ \\
    \bottomrule
    \end{tabularx}
\end{table}

\subsection{Training and Implementation Details}

Lifelong trajectory prediction encompasses sequential scenarios with dynamic data distributions as described in Section~\ref{Section-III}. 
Four groups of experiments with different numbers of lifelong learning tasks were constructed, as depicted in Table~\ref{table_datasets}. The number of tasks is $N \in \{5, 6, 7, 8\}$. The model is orderly trained in each group with the data stream from task $\text{T}_1$ to task $\text{T}_N$. All the training samples are fed to the model for one-time observation. After learning the $c$\textsuperscript{th} task, the model is tested with all the learned tasks from task $\text{T}_1$ to the current task $\text{T}_c$. Note that the task ID is used in the testing phase to distinguish different tasks for a clear and comprehensive evaluation. However, the task ID is unavailable for task-free CL methods during the training phase.

All the models are trained with a $1\times10^{-3}$ learning rate, and Adam is adopted as the optimizer~\cite{kingma2014adam}. The size of the training batch is set as 8. The coefficients $\alpha$ and $\beta$ in \eqref{eq_total_loss} are both 1. All the experiments are conducted on a Linux server with AMD EPYC-7763 CPU and NVIDIA GeForce RTX 4090 GPU. The average results and the standard deviations are obtained by repeating the experiments ten times. Between different repeated experiments, the training samples are shuffled within each task.

\subsection{Evaluation Metrics}\label{Section-V-B}
\subsubsection{Metrics for Prediction Accuracy}
Metrics used in \cite{li2024UQnet} are adopted to evaluate the trajectory prediction accuracy, including the minimum final displacement error (FDE) and the miss rate (MR). The UQnet derives $W$ predicted endpoints of TV from the heatmap $\hat{\mathbf{Y}}^{\text{hm}}$, where the $k$\textsuperscript{th} endpoint is denoted as ${\bf{\hat {\bf{Y}}}}_{k}$. FDE measures the minimum Euclidean distance between the $W$ predicted endpoints and the ground truth ${\bf{Y}}$ over each sample, which can be represented as:
\begin{equation}
    \text{FDE}^\text{sample} =\min_{k\in\left \{ 1,...,W \right \} }  {\left\| {{\bf{\hat {\bf{Y}}}}_{k} - {\bf{Y}}} \right\|}_2 \label{eq_fde_sample}
\end{equation}

Let $\text{FDE}^{\text{sample}}_{j}$ denote the FDE calculated on the $j$\textsuperscript{th} sample, and $N_{\text{T}}^{\text{test}}$ is the number of testing samples in task $\text{T}$. The FDE for the testing set of task $\text{T}$ is calculated as:
\begin{equation}
    \text{FDE}_{\text{T}}=\frac{1}{N_{\text{T}}^{\text{test}}}\sum_{j=1}^{N_{\text{T}}^{\text{test}}}\text{FDE}^{\text{sample}}_{j} \label{eq_fde_task}
\end{equation}

According to \eqref{eq_fde_task}, the average FDE over $N$ tasks $\{\text{T}_1,...,\text{T}_N\}$ can be calculated as:
\begin{equation}
    \text{FDE-AVG}=\frac{1}{N}\sum_{i=1}^{N}\text{FDE}_{\text{T}_i}
\end{equation}

MR measures the percentage of predicted endpoints that are out of a given lateral or longitudinal area of the ground truth. As implemented in \cite{li2024UQnet}, the lateral threshold of the given area is 1 $\rm{m}$, and the longitudinal threshold is determined by the piece-wise function depending on the velocity $v$ of the TV:

\begin{equation}
th_{\text{MR}}(v) = \left\{ \begin{array}{l}
1, v < 1.4 \\
1 + \frac{{v - 1.4}}{{11 - 1.4}}, 1.4 \le v \le 11\\
2, v > 11
\end{array} \right.
\end{equation}
where the unit of the longitudinal threshold is $\rm{m}$, and the unit of the velocity $v$ is $\rm{ms^{-1}}$. Let $W^{\text{out}}_{\text{T}}$ denote the number of predicted endpoints that are out of the given area in task $\text{T}$, the MR in the testing set of task $\text{T}$ is calculated as:
\begin{equation}
    \text{MR}_{\text{T}}= \frac{W^{\text{out}}_{\text{T}}}{N_{\text{T}}^{\text{test}} \times W}
\end{equation}

Furthermore, the average MR over $N$ tasks $\{\text{T}_1,...,\text{T}_N\}$ is calculated as:

\begin{equation}
    \text{MR-AVG} = \frac{1}{N} \sum_{i=1}^{N}\text{MR}_{\text{T}_i}
\end{equation}

Since FDE and MR both measure the prediction errors, the smaller values of FDE and MR represent the better prediction performance.

\subsubsection{Metrics for Catastrophic Forgetting}
For the evaluation of the capability to mitigate catastrophic forgetting, FDE-based backward transfer (FDE-BWT) and MR-based backward transfer (MR-BWT) are proposed to quantify catastrophic forgetting in lifelong trajectory prediction. If the prediction errors at task $\text{T}_j$ after training on task $\text{T}_i$ is denoted as $R_{i,j}$, FDE-BWT and MR-BWT can be computed uniformly by:
\begin{equation}
    \text{R-BWT}=\frac{1}{c-1} \sum_{i=1}^{c-1}\left (R_{c,i}-R_{i,i}  \right )  \label{eq-BWT}
\end{equation}
where $c \ge 2$ indexes the current task $\text{T}_c$. It measures the average error increment on all previously learned tasks after learning $\text{T}_c$. FDE-BWT and MR-BWT can be computed by replacing the error $R$ in \eqref{eq-BWT} as the values of FDE and MR, respectively. The smaller FDE-BWT and MR-BWT indicate better CL performance to avoid catastrophic forgetting.

\begin{figure*}[ht]
  \centering
  \includegraphics[scale=1.0]{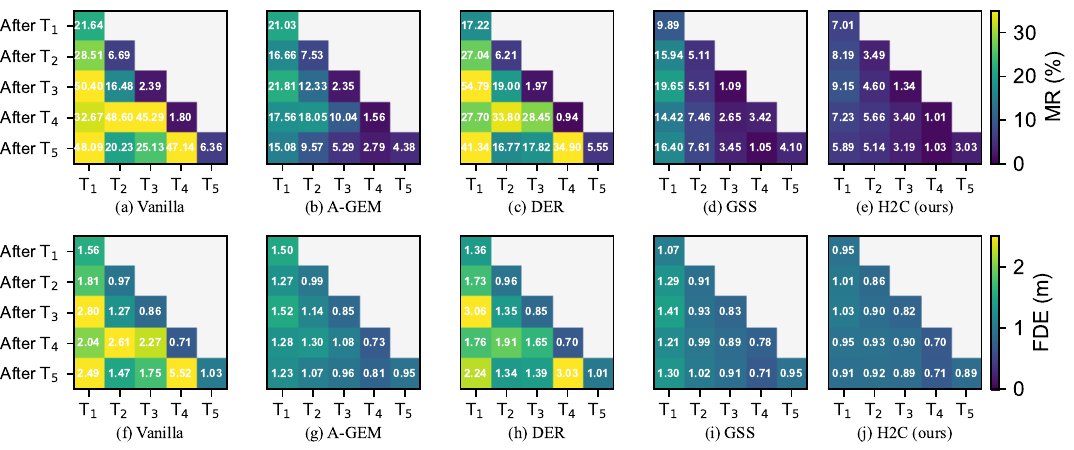}
  \caption{\justifying{MR and FDE calculated on every testing set of learned tasks in Group \uppercase\expandafter{\romannumeral1}. Within each sub-figure, the horizontal axis indexes the testing set, and the vertical axis indexes the latest learned task in sequential training. The buffer size is 2,000 for all CL methods.}}
  \label{array-mr}
\end{figure*}

\subsection{Baselines}
All the compared methods are applied to the UQnet. The detailed model settings are as follows:
\begin{itemize}
    \item \textbf{H2C (ours)}: The proposed task-free CL method.
    \item \textbf{Vanilla}: The UQnet without applying CL methods, acts as the non-CL baseline.
    \item \textbf{A-GEM}~\cite{chaudhry2019efficient-agem}: A task-based CL method that applies the inequality constraint to the average loss on all the replayed samples. The task ID is assumed to be available for this method.
    \item \textbf{DER}~\cite{buzzega2020dark}: A task-free CL method that uses reservoir sampling for the memory selection. It encourages the model to mimic the initial response for memory samples during the replay.
    \item \textbf{GSS}~\cite{aljundi2019-gss}: A task-free CL method that uses a gradient-based sampling for the memory selection. It trains the model with the mixed samples from the current task and the memory buffer.
    \item \textbf{Joint}: The UQnet with joint training. All the training data are available simultaneously in each experimental group, which does not follow the assumption of CL.
\end{itemize}

The buffer size refers to the maximum number of memory samples that can be stored for each CL method. According to this definition, the buffer size for H2C is the summation of $|\mathcal{M}_\text{sp}|$ and $|\mathcal{M}_\text{cp}|$, and we set that $|\mathcal{M}_\text{sp}|$ is equal to $|\mathcal{M}_\text{cp}|$ in the experiments. All the compared CL methods are assumed to have the same buffer size. Moreover, only the Joint learns with an enormous dataset that covers all the data in all tasks. Since the data stream is one-time observed for CL methods, the training epoch is 1 for the Joint for a fair comparison.

\begin{table*}[ht]
\centering
\caption{FDE-AVG (m) and FDE-BWT (m) with the buffer size 2,000 for CL methods. Joint learns with all the data in a group instead of learning with the data stream continually. Thus, FDE-BWT is not applicable for Joint, which is denoted as N/A.}
\resizebox{\linewidth}{!}{
    \begin{tabular}{c|cc|cc|cc|cc}
    \toprule
   Task group & \multicolumn{2}{c|}{Group \uppercase\expandafter{\romannumeral1}: task $\text{T}_1$ - task $\text{T}_5$} & \multicolumn{2}{c|}{Group \uppercase\expandafter{\romannumeral2}: task $\text{T}_1$ - task $\text{T}_6$} & \multicolumn{2}{c|}{Group \uppercase\expandafter{\romannumeral3}: task $\text{T}_1$ - task $\text{T}_7$} & \multicolumn{2}{c}{Group \uppercase\expandafter{\romannumeral4}: task $\text{T}_1$ - task $\text{T}_8$} \\
    \midrule
   Evaluation metrics    & FDE-AVG ($\downarrow$) & FDE-BWT ($\downarrow$) & FDE-AVG ($\downarrow$) & FDE-BWT ($\downarrow$) & FDE-AVG ($\downarrow$) & FDE-BWT ($\downarrow$) & FDE-AVG ($\downarrow$) & FDE-BWT ($\downarrow$) \\
    \midrule
    Vanilla    & 2.45 $\pm$ 0.27 & 1.78 $\pm$ 0.41 & 2.23 $\pm$ 0.35 & 1.51 $\pm$ 0.42 & 1.51 $\pm$ 0.16 & 0.66 $\pm$ 0.13 & 2.35 $\pm$ 0.20  & 1.65 $\pm$ 0.24 \\
    A-GEM~\cite{chaudhry2019efficient-agem}      & 1.01 $\pm$ 0.60 & 0.12 $\pm$ 0.11 & 1.09 $\pm$ 0.10 &  0.20 $\pm$ 0.11 & 1.00 $\pm$ 0.08 & 0.12 $\pm$ 0.09 & 1.01 $\pm$ 0.10 & 0.18 $\pm$ 0.12 \\
    
    DER~\cite{buzzega2020dark}        & 1.80 $\pm$ 0.32 & 1.03 $\pm$ 0.40 & 1.40 $\pm$ 0.24 & 0.56 $\pm$ 0.24 & 1.34 $\pm$ 0.23 & 0.51 $\pm$ 0.24 & 1.33 $\pm$ 0.11 & 0.52 $\pm$ 0.11 \\
    GSS~\cite{aljundi2019-gss}        & 0.98 $\pm$ 0.06 & 0.10 $\pm$ 0.08 & 0.94 $\pm$ 0.07 & 0.09 $\pm$ 0.06 & 0.99 $\pm$ 0.07 & 0.16 $\pm$ 0.11 & 1.12 $\pm$ 0.13 & 0.32 $\pm$ 0.16 \\
    H2C (ours)       & \textbf{0.86 $\pm$ 0.03} & \textbf{0.04 $\pm$ 0.02} & \textbf{0.89 $\pm$ 0.05} & \textbf{0.08 $\pm$ 0.07} & \textbf{0.89 $\pm$ 0.02} & \textbf{0.09 $\pm$ 0.03} & 0.91 $\pm$ 0.03 & \textbf{0.13 $\pm$ 0.03} \\
    Joint      & 0.98 $\pm$ 0.07 &       N/A       & 1.00 $\pm$ 0.15 &       N/A       & 0.90 $\pm$ 0.07 &       N/A       & \textbf{0.84 $\pm$ 0.07} &       N/A       \\
    \bottomrule
    \end{tabular}
}\label{table-overall-fde}
\end{table*}

\begin{table*}[ht]
\centering
\caption{MR-AVG (\%) and MR-BWT (\%) with the buffer size 2,000 for CL methods. Joint learns with all the data in a group instead of learning with the data stream continually. Thus, MR-BWT is not applicable for Joint, which is denoted as N/A.}
\resizebox{\linewidth}{!}{

    \begin{tabular}{c|cc|cc|cc|cc}
    \toprule
   Task group & \multicolumn{2}{c|}{Group \uppercase\expandafter{\romannumeral1}: task $\text{T}_1$ - task $\text{T}_5$} & \multicolumn{2}{c|}{Group \uppercase\expandafter{\romannumeral2}: task $\text{T}_1$ - task $\text{T}_6$} & \multicolumn{2}{c|}{Group \uppercase\expandafter{\romannumeral3}: task $\text{T}_1$ - task $\text{T}_7$} & \multicolumn{2}{c}{Group \uppercase\expandafter{\romannumeral4}: task $\text{T}_1$ - task $\text{T}_8$} \\
    \midrule
   Evaluation metrics    & MR-AVG ($\downarrow$) & MR-BWT ($\downarrow$) & MR-AVG ($\downarrow$) & MR-BWT ($\downarrow$) & MR-AVG ($\downarrow$) & MR-BWT ($\downarrow$) & MR-AVG ($\downarrow$) & MR-BWT ($\downarrow$) \\
    \midrule
    Vanilla    & 29.39 $\pm$ 3.69 & 27.02 $\pm$ 5.84 & 27.55 $\pm$ 7.07 & 25.01 $\pm$ 9.20 & 19.82 $\pm$ 3.19 & 15.69 $\pm$ 2.90 & 32.54 $\pm$ 2.52 & 30.81 $\pm$ 3.15 \\
    A-GEM~\cite{chaudhry2019efficient-agem}      & 7.42 $\pm$ 1.57 & 3.06 $\pm$ 2.75 & 10.34 $\pm$ 3.14 & 5.72 $\pm$ 3.74 & 7.47 $\pm$ 2.57 & 3.21 $\pm$ 2.64 &8.75 $\pm$ 2.74 & 4.08 $\pm$ 4.72 \\
    DER~\cite{buzzega2020dark}        & 23.27 $\pm$ 4.88 & 21.12 $\pm$ 6.53 & 17.61 $\pm$ 3.39 & 14.50 $\pm$ 3.13 & 15.73 $\pm$ 4.88 & 12.12 $\pm$ 4.92 & 17.09 $\pm$ 2.98 & 14.18 $\pm$ 3.14 \\
    GSS~\cite{aljundi2019-gss}        &  6.52 $\pm$ 1.38 & 2.56 $\pm$ 2.22 &  5.41 $\pm$ 1.95  & 2.18 $\pm$ 1.44 &  7.17 $\pm$ 2.13 & 3.79 $\pm$ 3.36 & 10.44 $\pm$ 2.86 & 7.97 $\pm$ 3.87 \\
    H2C (ours)       &  \textbf{3.66 $\pm$ 1.04} & \textbf{1.02 $\pm$ 1.18} &  \textbf{4.22 $\pm$ 1.52} & \textbf{1.70 $\pm$ 1.92} &  \textbf{4.50 $\pm$ 0.40} & \textbf{1.97 $\pm$ 0.72} & 5.10 $\pm$ 0.58 & \textbf{3.02 $\pm$ 0.76} \\
    Joint      &  6.31 $\pm$ 2.16 &        N/A      &  6.42 $\pm$ 2.50 &       N/A       &  5.40 $\pm$ 2.14 &       N/A       & \textbf{4.14 $\pm$ 1.83} &       N/A       \\
    \bottomrule
    \end{tabular}
}\label{table-overall-mr}
\end{table*}

\subsection{Experimental Results with Analysis}\label{Section-V-D}
\subsubsection{Catastrophic Forgetting in Lifelong Trajectory Prediction}
Catastrophic forgetting refers to the significant degradation of the prediction accuracy in previously learned tasks after the model adapts to a new task. To investigate this phenomenon, the models are tested in all the learned tasks when finishing a new one. Fig.~\ref{array-mr} demonstrates the MR and FDE in Group \uppercase\expandafter{\romannumeral1} with five tasks. Within each sub-figure in  Fig.~\ref{array-mr}, the horizontal axis indexes the testing set. The vertical axis indexes the latest task learned by the model, which refers to the current task in the training process. For a convenience of description, we denote the MR and FDE at the $i$\textsuperscript{th} row and the $j$\textsuperscript{th} column in each sub-figure of Fig.~\ref{array-mr} as $\text{MR}_{i,j}$ and $\text{FDE}_{i,j}$, respectively. In other words, $\text{MR}_{i,j}$ and $\text{FDE}_{i,j}$ demonstrate the test performance in task $\text{T}_j$ after training in $\text{T}_i$.

First, we focus on the performance of the Vanilla, as shown in Fig.~\ref{array-mr}(a) and Fig.~\ref{array-mr}(f). Comparing the MR in each row, it can be found that $\forall i >j$, $\text{MR}_{i, j} >\text{MR}_{i,i}$. Similar results are found in Fig.~\ref{array-mr}(f), where $\forall i >j$, $\text{FDE}_{i, j} > \text{FDE}_{i,i}$. These results show that Vanilla achieves more accurate prediction in the current task than previously learned ones in such continuous training. It suggests that the model parameters tend to adapt to more recent samples from the data stream. Furthermore, the catastrophic forgetting of Vanilla is demonstrated by comparisons among columns in Fig.~\ref{array-mr}(a) and (f), where the MR and FDE in the same testing set increase significantly after learning new tasks. For example, $\text{MR}_{1,1}$ is 21.64\% in the testing set of $\text{T}_1$ when the model is trained with $\text{T}_1$. However, the $\text{MR}_{2,1}$ tested in $\text{T}_1$ becomes 28.51\% after the model learns $\text{T}_2$. Moreover, the MR becomes 50.4\% after the model sequentially learns new tasks $\{\text{T}_2,\text{T}_3\}$. Comparing $\text{MR}_{3,1}$ and $\text{MR}_{1,1}$, the MR is increased by 28.76\%, revealing the catastrophic forgetting of Vanilla. 

\begin{figure*}[htbp]
\centering
\begin{subfigure}[b]{1\textwidth} 
\centering
\includegraphics[scale=1]{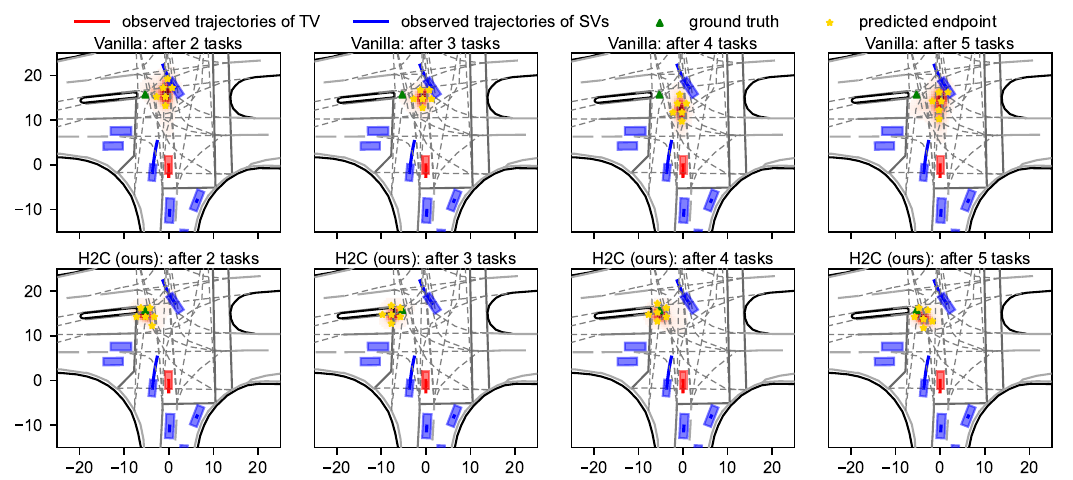}
\caption{A test case from task $\text{T}_1$.}
\label{subfig-vis-ma}
\end{subfigure}

\begin{subfigure}[b]{1\textwidth} 
\centering
\includegraphics[scale=1]{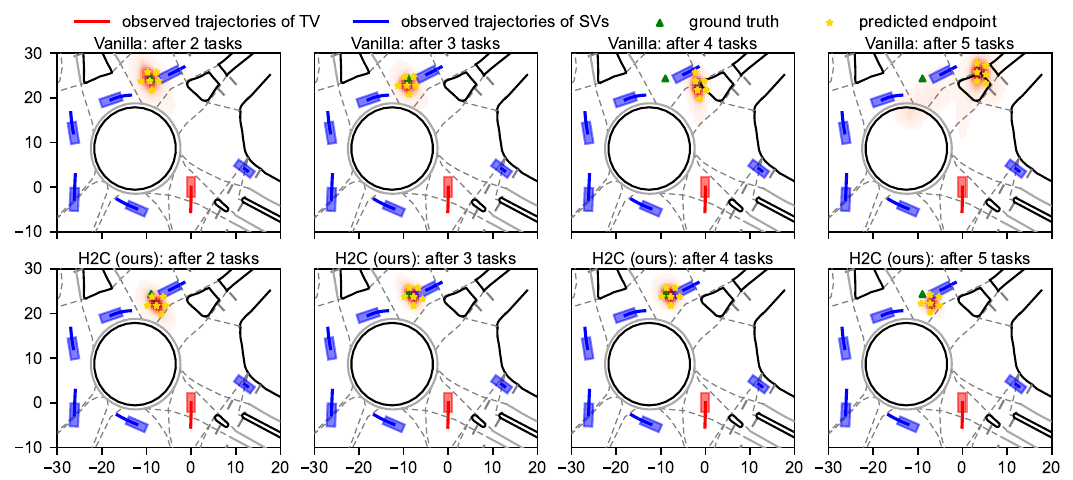}
\caption{A test case from task $\text{T}_2$.}
\label{subfig-vis-ft}
\end{subfigure}
    \caption{Two test cases comparing Vanilla and H2C. TV and SVs are colored by red and blue, respectively. The ground truth is the endpoint of the TV at the future 3s, denoted as the green triangle. The predicted endpoints sampled from the output heatmap are denoted as yellow stars.}
    \label{fig-case-vis}
\end{figure*}

Comparing each column in Fig.~\ref{array-mr}(a) with Fig.~\ref{array-mr}(e), the proposed H2C has smaller increments between $\text{MR}_{j,j}$ and $\text{MR}_{i,j}$ than Vanilla when $i > j$. H2C also reduces the FDE increment of Vanilla, as shown in the comparison between Fig.~\ref{array-mr}(f) and Fig.~\ref{array-mr}(j). For a more intuitive comparison, a test case in $\text{T}_1$ and a test case in $\text{T}_2$ are visualized as depicted in Fig.~\ref{fig-case-vis}(a) and Fig.~\ref{fig-case-vis}(b), respectively. The scenario of $\text{T}_1$ is an intersection. In the test case in task $\text{T}_1$, the predicted endpoints and the range of heatmaps drift away from the ground truth after Vanilla learns 4 and 5 tasks. Compared to Vanilla, the predicted endpoints from H2C are closer to the ground truth. Task $\text{T}_2$ demonstrate a roundabout scenario, H2C retains the accurate prediction in this test case of $\text{T}_2$ after learning tasks including $\{\text{T}_2, \text{T}_3, \text{T}_4, \text{T}_5\}$. Vanilla obtains an accurate prediction after learning $\text{T}_2$ and $\text{T}_3$. However, the distance between the prediction and the ground truth becomes large when learning more tasks. These experimental results demonstrate that the proposed H2C alleviates catastrophic forgetting of the non-CL model. As depicted in Fig.~\ref{array-mr}(a)-(d) and Fig.~\ref{array-mr}(f)-(i), other CL methods have smaller error increments than Vanilla, indicating that catastrophic forgetting of Vanilla is also mitigated by these CL methods. 


\begin{figure*}[ht]
  \centering
  \includegraphics[scale=0.9]{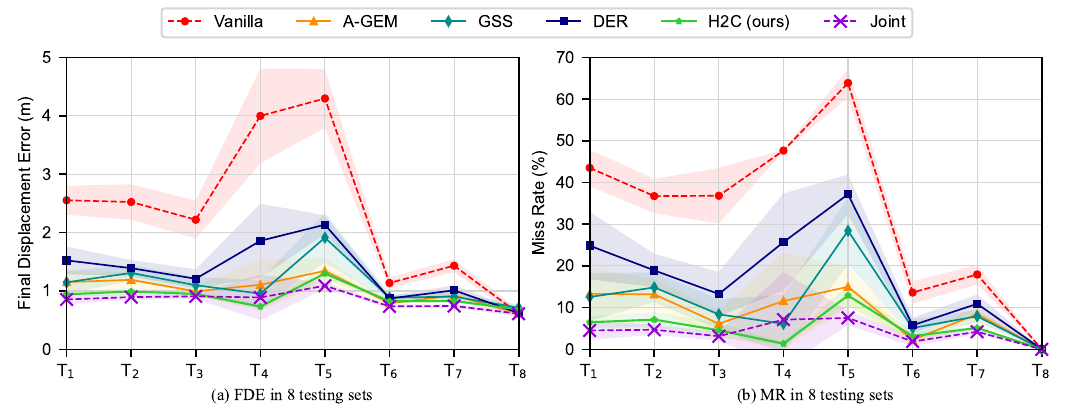}
  \caption{\justifying{Detailed performance measured by FDE (a) and MR (b) in 8 lifelong learning tasks. The shaded region represents the standard deviation, displayed as a translucent color band around the value. Joint is trained with a dataset built by all the training data from 8 tasks, and tested in each task, respectively. All the CL methods and Vanilla are continually trained with the data stream from task 1 to task 8, and tested after finishing the 8\textsuperscript{th} task. The buffer size for CL methods is 2,000.}}
  \label{linechart}
\end{figure*}

\subsubsection{Overall Performance}
To further validate the effectiveness of H2C, more experimental groups were conducted. As shown in Table~\ref{table-overall-fde} and Table~\ref{table-overall-mr}, FDE-BWT and MR-BWT are used to measure the capability to mitigate catastrophic forgetting, and the overall performance of prediction accuracy is evaluated by FDE-AVG and MR-AVG. Compared to Vanilla, FDE-BWT and MR-BWT are smaller in CL methods in most cases. Among four groups, the proposed H2C mitigates catastrophic forgetting of Vanilla by an average 22.71\% reduction of MR-BWT, and the average reduction of FDE-BWT is 5.26 m. Moreover, H2C has the lowest FDE-BWT and MR-BWT among compared CL methods, which shows the advantage in mitigating catastrophic forgetting. From the aspect of overall performance, Vanilla has the maximum FDE-AVG and MR-AVG in most cases. H2C also achieves the lowest FDE-AVG and MR-AVG in compared CL methods.

Furthermore, an important motivation of using CL to handle lifelong tasks is improving the efficiency of joint training~\cite{TPAMI2024-comprehensive-cl}. The strategy of joint training assumes that data from all potentially encountered tasks is available at once. However, it may be impractical to a due to privacy or safety concerns~\cite{li2021privacy}. Meanwhile, storing and maintaining a vast amount of data from potentially unlimited tasks may overwhelm the storage and computation resources in AD system. Differently, CL methods aim to learn from a sequentially available data stream. Under this assumption, CL methods are expected to achieve similar or better performance compared to the strategy of joint training. In our experiments, the model applied with joint training is denoted as Joint. As shown in Table~\ref{table-overall-fde} and Table~\ref{table-overall-mr}, Joint outperforms most CL methods, including A-GEM, GSS, and DER, by significantly lower FDE-AVG and MR-AVG in most cases. For example, in Group~\uppercase\expandafter{\romannumeral3}, the gaps between Joint and DER are 0.44 m for FDE-AVG, and 10.33\% for MR-AVG. In Group~\uppercase\expandafter{\romannumeral4}, the gaps between Joint and DER are 0.49 m for FDE-AVG and 12.95\%. Nevertheless, the proposed H2C outperforms Joint by lower FDE-AVG and MR-AVG in Group \uppercase\expandafter{\romannumeral1} to Group \uppercase\expandafter{\romannumeral3}. The performance gap between H2C and Joint is small in Group \uppercase\expandafter{\romannumeral4}, where Joint has a lower FDE-AVG than H2C by 0.07 m and by 0.96\% for MR-AVG.

\subsubsection{Performance Stability}
The standard deviations in Table~\ref{table-overall-fde} and Table~\ref{table-overall-mr} measure the amount of variation of results among repeated 10 times experiments, demonstrating the influence of shuffled samples on the performance stability. It can be found that the standard deviations of all the metrics from H2C are the minimum in most cases.

To compare the performance stability in varying tasks, detailed FDE and MR tested in every task from Group~\uppercase\expandafter{\romannumeral4} are depicted in Fig.~\ref{linechart}. The experimental results of Vanilla and Joint are used as references with dotted lines, marked by red dots and purple crosses, respectively. It can be found that Vanilla has a larger variation of FDE and MR than Joint. For example, the tested FDE and MR of Vanilla and Joint in task $\text{T}_8$ are close. Meanwhile, the maximum FDE and MR appear in the test of task $\text{T}_5$, both for Vanilla and Joint. However, in task $\text{T}_5$, the FDE of Vanilla is more than 4 m, and the FDE of Joint is approximately 1 m. Compared with Vanilla and Joint, the performance of H2C is close to Joint. Although other CL methods, including A-GEM, DER, and GSS, have lower FDE and MR than Vanilla in most tested tasks, they also have relatively large variations in these tasks. These experimental results demonstrate the advance of H2C in keeping accurate and stable predictions.


\subsubsection{Influence of Buffer Size on Continual Learning}
In the experiments, all of the compared CL methods use memory buffers to store samples for replay-based CL strategies. The buffer size is a nontrivial hyper-parameter for replay-based CL~\cite{lesort2020cl-robot-survey}. To investigate the influence of the buffer size on the performance of CL, the FDE-BWT and MR-BWT of CL methods applied with different buffer sizes are compared. Since the buffer size is required to be far less than the total number of training samples, as the constraint formulated in \eqref{eq_amount}, we employ four settings of the buffer size, including 500, 1,000, 2,000, and 4,000 batches in the experiments. 

As shown in Table~\ref{table_buffer}, the FDE-BWT in Group~\uppercase\expandafter{\romannumeral4} with four different buffer size settings are compared. Let $|\mathcal{M}|$ denote the buffer size, and the total amount of training samples in the lifelong tasks is denoted as $N_\text{total}^{\text{train}} = \sum_{i=1}^{N}N_{\text{T}_i}$, the ratio in Table~\ref{table_buffer} is calculated by $\text{Ratio}=|\mathcal{M}|/N_\text{total}^{\text{train}}$. We can find that H2C has the minimum FDE-BWT compared to other CL methods among these four settings of batch size. Moreover, the FDE-BWT of H2C and GSS decreased with the increment of buffer size in these four settings. A similar relationship between the buffer size and FDE-BWT also occurs in A-GEM with 500, 1,000, and 2,000 buffer sizes. However, the changing of buffer size has little impact on DER.

\begin{figure}[tp]
  \centering

  \includegraphics[scale=1.0]{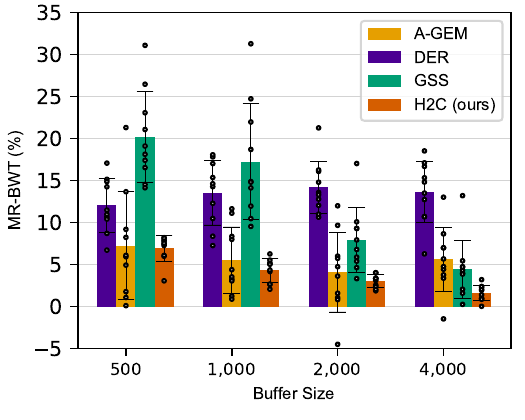}
  \caption{\justifying{The MR-BWT compared between CL methods with four settings of the buffer size.}}
  \label{barchart}
\end{figure}

Experimental results for MR-BWT are depicted in Fig.~\ref{barchart}. The error bar in Fig.~\ref{barchart} is obtained by repeating experiments for 10 times, and the scatters are detailed values of MR-BWT in these repeated experiments. Similar to the results shown in Table~\ref{table_buffer}, the proposed H2C has the lowest MR-BWT in all settings. Note that the MR-BWT of GSS are higher than DER and A-GEM when the buffer size is 500 and 1,000. Interestingly, the performance of GSS surpasses DER and A-GEM with the lower MR-BWT when the buffer size is 4,000. These experimental results show that the performance of H2C and GSS are sensitive to the buffer size, where the capability of CL can be enhanced by increasing the amount of replayed samples. Moreover, the smaller error deviations and the shorter error bars also indicate that H2C has a more stable performance in 10 repeated experiments.

\begin{table}[tp]
    \centering
    \caption{FDE-BWT (m) in Group~\uppercase\expandafter{\romannumeral4} (8 tasks) with different buffer Sizes. The ratio refers to the proportion of buffer size to the total number of training samples.}
    \label{table_buffer}
    \begin{tabularx}{\linewidth}{c | *{4}{>{\centering\arraybackslash}X}}
    \toprule
     \makecell{Size\\(Ratio)}  & \makecell{500\\(0.20\%)} &\makecell{1,000\\(0.40\%)} & \makecell{2,000\\(0.80\%)} & \makecell{4,000\\(1.60\%)}  \\ 
    \midrule 
    A-GEM   &  0.30 $\pm$ 0.29 &  0.20 $\pm$ 0.14 & 0.14  $\pm$ 0.17 & 0.22 $\pm$  0.13 \\
    DER     & 0.46 $\pm$ 0.14 & 0.54 $\pm$ 0.19 & 0.52 $\pm$ 0.11 & 0.51 $\pm$ 0.17 \\
    GSS     & 0.86 $\pm$ 0.23 & 0.77 $\pm$ 0.39 & 0.32 $\pm$ 0.16 & 0.19 $\pm$ 0.14 \\
    H2C       & \textbf{0.27 $\pm$ 0.06} & \textbf{0.18 $\pm$ 0.05} & \textbf{0.13 $\pm$ 0.03} & \textbf{0.08 $\pm$ 0.03}   \\
    \bottomrule
    \end{tabularx}
\end{table}

\subsection{Discussion}
Coming back to the research question presented in Section~\ref{Section-I}, this study focuses on developing a novel CL method that can effectively retain learned knowledge without accessing the task boundary. As an answer, we propose H2C to selectively replay learned samples in a task-free manner.

Comparisons between Vanilla and H2C in Section~\ref{Section-V} have demonstrated the capability of H2C to mitigate catastrophic forgetting of the DNN-based model without accessing the task boundary. In the discussion, we dive deeper into the experimental comparison between H2C and Joint to further reveal the resource efficiency of H2C. Joint is trained using all samples simultaneously. CL studies usually consider such joint training as the upper bound of performance~\cite{bao2023lifelong}. Among all CL methods in the experiments, H2C achieves the closest performance to Joint. Referred to Table~\ref{table_datasets}, Joint requires using 252,704 samples from 8 scenarios in Group~\uppercase\expandafter{\romannumeral4}, while the storage cost of the memory buffer from H2C is only 0.80\% of Joint when the buffer size is 2,000. In such cases, the FDE-AVG is 0.84 m for Joint and 0.91 m for H2C, as depicted in Table~\ref{table-overall-fde}. Moreover, as the analysis presented in Section~\ref{Section-V-D}, H2C outperforms Joint in Group~\uppercase\expandafter{\romannumeral1}, Group~\uppercase\expandafter{\romannumeral2}, and Group~\uppercase\expandafter{\romannumeral3}. These experimental results show that H2C can achieve comparable or superior performance to the joint training approach while requiring significantly fewer computational resources.

To further validate the effectiveness of retaining learned knowledge, we discuss the performance of compared task-free CL methods. DER~\cite{buzzega2020dark} is a representative task-free CL method~\cite{TPAMI2024-comprehensive-cl}. Taking DER for example, it models an overall representation of learned knowledge by random sampling~\cite{buzzega2020dark}. However, it may cause an imbalanced resource allocation when the data amount significantly varies between tasks. For instance, task $\text{T}_1$ has 33,456 training samples while task $\text{T}_2$ has 66,256 training samples, referred to Table~\ref{table_datasets}. The ratio of training data amount is approximately 1:2. Using random sampling, the ratio of stored memory samples for $\text{T}_1$ and $\text{T}_2$ is nearly the same as the ratio of those training samples after DER learns $\{\text{T}_1, \text{T}_2\}$. These imbalanced memory samples lead to an imbalanced replay, where the memory resources allocated for $\text{T}_1$ is far less than those for $\text{T}_2$.

We compare the error increments in $\text{T}_1$ and $\text{T}_2$ to investigate the impact of the imbalanced replay in DER. As depicted in Fig.~\ref{array-mr}(c), $\text{MR}_{1,1}$ is 17.22\%, and $\text{MR}_{2,2}$ is 6.21\%. After DER learned three tasks in the experiments, $\text{MR}_{3,1}$ is 54.79\%, while $\text{MR}_{3,2}$ is 19.00\%. The increment of MR in $\text{T}_1$ is 37.57\%, calculated by $\text{MR}_{3,1}-\text{MR}_{1,1}$. However, MR only increases 12.79\% in $\text{T}_2$. The imbalanced replaying between $\text{T}_1$ and $\text{T}_2$ could be an important factor to such CL performance gap. To address the problem of imbalanced replay, the proposed H2C integrates distinctive and overall representations via the HPC-inspired complementary strategies. Referred to the same instance in Fig.~\ref{array-mr}(e), MR increment in $\text{T}_1$ after H2C learned three tasks is 0.96\%, and MR increment in $\text{T}_2$ is 1.11\%. Experimental results including the less performance gap and the lower MR demonstrate that H2C achieves better capability to retain learned knowledge than compared task-free CL methods.

\section{Conclusion and Future Work}\label{Section-VI}

This study proposes a novel task-free CL method, H2C, to mitigate catastrophic forgetting in lifelong trajectory prediction without relying on task boundaries. Drawing inspiration from HPC, the proposed H2C constructs the separation buffer and the completion buffer to represent learned knowledge via two strategies. The separation buffer captures the distinctive knowledge by maximizing inter-sample diversity while the completion buffer employs an equiprobable sampling strategy to estimate the overall knowledge. Replaying such complementary knowledge enhances the applicability of DNN-based trajectory prediction model in real-world AD. Experimental results demonstrate that H2C significantly mitigates catastrophic forgetting in DNN-based trajectory prediction models, with the mean reduction of the MR-BWT by 22.71\% and the FDE-BWT by 5.26 m on average.

In future work, we will investigate how the lifelong trajectory prediction performance achieved by H2C influences the motion planning in lifelong AD tasks. In addition, exploring the integration of CL with downstream modules in AD, such as decision-making and motion planning, is also meaningful for developing high-level AD and building safe and efficient intelligent transportation systems.

\small
\bibliographystyle{IEEEtran}
\bibliography{ref.bib}

\begin{thebibliography}{10}
\providecommand{\url}[1]{#1}
\csname url@samestyle\endcsname
\providecommand{\newblock}{\relax}
\providecommand{\bibinfo}[2]{#2}
\providecommand{\BIBentrySTDinterwordspacing}{\spaceskip=0pt\relax}
\providecommand{\BIBentryALTinterwordstretchfactor}{4}
\providecommand{\BIBentryALTinterwordspacing}{\spaceskip=\fontdimen2\font plus
\BIBentryALTinterwordstretchfactor\fontdimen3\font minus \fontdimen4\font\relax}
\providecommand{\BIBforeignlanguage}[2]{{%
\expandafter\ifx\csname l@#1\endcsname\relax
\typeout{** WARNING: IEEEtran.bst: No hyphenation pattern has been}%
\typeout{** loaded for the language `#1'. Using the pattern for}%
\typeout{** the default language instead.}%
\else
\language=\csname l@#1\endcsname
\fi
#2}}
\providecommand{\BIBdecl}{\relax}
\BIBdecl

\bibitem{pred-plan2025-its}
W.~Shao, J.~Xu, Z.~Cao, H.~Wang, and J.~Li, ``From prediction to planning: Comprehensive uncertainty management in autonomous driving,'' \emph{IEEE Transactions on Intelligent Transportation Systems}, pp. 1--15, 2025.

\bibitem{diffusion2025-its}
Z.~Lan, L.~Liu, Y.~Ren, Z.~Cui, and H.~Yu, ``Diffutory: A future feature and mode association augmented diffusion model for trajectory prediction in autonomous vehicles,'' \emph{IEEE Transactions on Intelligent Transportation Systems}, pp. 1--16, 2025.

\bibitem{mozaffari2020deep}
S.~Mozaffari, O.~Y. Al-Jarrah, M.~Dianati, P.~Jennings, and A.~Mouzakitis, ``Deep learning-based vehicle behavior prediction for autonomous driving applications: A review,'' \emph{IEEE Transactions on Intelligent Transportation Systems}, vol.~23, no.~1, pp. 33--47, 2022.

\bibitem{kudithipudi2022biological-nature}
D.~Kudithipudi, M.~Aguilar-Simon, J.~Babb, M.~Bazhenov, D.~Blackiston, J.~Bongard, A.~P. Brna, S.~Chakravarthi~Raja, N.~Cheney, J.~Clune \emph{et~al.}, ``Biological underpinnings for lifelong learning machines,'' \emph{Nature Machine Intelligence}, vol.~4, no.~3, pp. 196--210, 2022.

\bibitem{lesort2020cl-robot-survey}
T.~Lesort, V.~Lomonaco, A.~Stoian, D.~Maltoni, D.~Filliat, and N.~D{\'\i}az-Rodr{\'\i}guez, ``Continual learning for robotics: Definition, framework, learning strategies, opportunities and challenges,'' \emph{Information fusion}, vol.~58, pp. 52--68, 2020.

\bibitem{forget}
R.~M. French, ``Catastrophic forgetting in connectionist networks,'' \emph{Trends in cognitive sciences}, vol.~3, no.~4, pp. 128--135, 1999.

\bibitem{2025Haochen-TPAMI}
H.~Liu, Z.~Huang, W.~Huang, H.~Yang, X.~Mo, and C.~Lv, ``Hybrid-prediction integrated planning for autonomous driving,'' \emph{IEEE Transactions on Pattern Analysis and Machine Intelligence}, pp. 1--18, 2025.

\bibitem{chen2018lifelong}
Z.~Chen and B.~Liu, \emph{Lifelong machine learning}.\hskip 1em plus 0.5em minus 0.4em\relax Springer, 2018, vol.~1.

\bibitem{van2022three-incremental}
G.~M. Van~de Ven, T.~Tuytelaars, and A.~S. Tolias, ``Three types of incremental learning,'' \emph{Nature Machine Intelligence}, vol.~4, no.~12, pp. 1185--1197, 2022.

\bibitem{TPAMI2024-comprehensive-cl}
L.~Wang, X.~Zhang, H.~Su, and J.~Zhu, ``A comprehensive survey of continual learning: Theory, method and application,'' \emph{IEEE Transactions on Pattern Analysis and Machine Intelligence}, vol.~46, no.~8, pp. 5362--5383, 2024.

\bibitem{wang2023incorporating-nmi}
L.~Wang, X.~Zhang, Q.~Li, M.~Zhang, H.~Su, J.~Zhu, and Y.~Zhong, ``Incorporating neuro-inspired adaptability for continual learning in artificial intelligence,'' \emph{Nature Machine Intelligence}, vol.~5, no.~12, pp. 1356--1368, 2023.

\bibitem{yin2023bioslamTRO}
P.~Yin, A.~Abuduweili, S.~Zhao, L.~Xu, C.~Liu, and S.~Scherer, ``Bioslam: A bioinspired lifelong memory system for general place recognition,'' \emph{IEEE Transactions on Robotics}, vol.~39, no.~6, pp. 4855--4874, 2023.

\bibitem{ma2021continual}
H.~Ma, Y.~Sun, J.~Li, M.~Tomizuka, and C.~Choi, ``Continual multi-agent interaction behavior prediction with conditional generative memory,'' \emph{IEEE Robotics and Automation Letters}, vol.~6, no.~4, pp. 8410--8417, 2021.

\bibitem{bao2023lifelong}
P.~Bao, Z.~Chen, J.~Wang, D.~Dai, and H.~Zhao, ``Lifelong vehicle trajectory prediction framework based on generative replay,'' \emph{IEEE Transactions on Intelligent Transportation Systems}, vol.~24, no.~12, pp. 13\,729--13\,741, 2023.

\bibitem{lin2024continual}
Y.~Lin, Z.~Li, C.~Gong, C.~Lu, X.~Wang, and J.~Gong, ``Continual interactive behavior learning with traffic divergence measurement: A dynamic gradient scenario memory approach,'' \emph{IEEE Transactions on Intelligent Transportation Systems}, vol.~25, no.~3, pp. 2355--2372, 2024.

\bibitem{zador2023catalyzing-neuroai}
A.~Zador, S.~Escola, B.~Richards, B.~{\"O}lveczky, Y.~Bengio, K.~Boahen, M.~Botvinick, D.~Chklovskii, A.~Churchland, C.~Clopath \emph{et~al.}, ``Catalyzing next-generation artificial intelligence through neuroai,'' \emph{Nature communications}, vol.~14, no.~1, p. 1597, 2023.

\bibitem{sun2025pattern}
L.~Sun, S.~Li, P.~Ren, Q.~Liu, Z.~Li, and X.~Liang, ``Pattern separation and pattern completion within the hippocampal circuit during naturalistic stimuli,'' \emph{Human Brain Mapping}, vol.~46, no.~2, p. e70150, 2025.

\bibitem{zeno2021task-agnostic}
C.~Zeno, I.~Golan, E.~Hoffer, and D.~Soudry, ``Task-agnostic continual learning using online variational bayes with fixed-point updates,'' \emph{Neural Computation}, vol.~33, no.~11, pp. 3139--3177, 2021.

\bibitem{zhu2024tame}
H.~Zhu, M.~Majzoubi, A.~Jain, and A.~Choromanska, ``Tame: Task agnostic continual learning using multiple experts,'' in \emph{Proceedings of the IEEE/CVF Conference on Computer Vision and Pattern Recognition}, 2024, pp. 4139--4148.

\bibitem{buzzega2020dark}
P.~Buzzega, M.~Boschini, A.~Porrello, D.~Abati, and S.~Calderara, ``Dark experience for general continual learning: a strong, simple baseline,'' \emph{Advances in neural information processing systems}, vol.~33, pp. 15\,920--15\,930, 2020.

\bibitem{bakker2008pattern-science}
A.~Bakker, C.~B. Kirwan, M.~Miller, and C.~E. Stark, ``Pattern separation in the human hippocampal ca3 and dentate gyrus,'' \emph{science}, vol. 319, no. 5870, pp. 1640--1642, 2008.

\bibitem{korbmacher2022deep-review}
R.~Korbmacher and A.~Tordeux, ``Review of pedestrian trajectory prediction methods: Comparing deep learning and knowledge-based approaches,'' \emph{IEEE Transactions on Intelligent Transportation Systems}, vol.~23, no.~12, pp. 24\,126--24\,144, 2022.

\bibitem{2016social-lstm}
A.~Alahi, K.~Goel, V.~Ramanathan, A.~Robicquet, L.~Fei-Fei, and S.~Savarese, ``Social lstm: Human trajectory prediction in crowded spaces,'' in \emph{Proceedings of the IEEE conference on computer vision and pattern recognition}, 2016, pp. 961--971.

\bibitem{2017RNN}
F.~Altché and A.~de~La~Fortelle, ``An lstm network for highway trajectory prediction,'' in \emph{2017 IEEE 20th International Conference on Intelligent Transportation Systems (ITSC)}, 2017, pp. 353--359.

\bibitem{Social-aware2022-pami}
P.~Zhang, J.~Xue, P.~Zhang, N.~Zheng, and W.~Ouyang, ``Social-aware pedestrian trajectory prediction via states refinement lstm,'' \emph{IEEE Transactions on Pattern Analysis and Machine Intelligence}, vol.~44, no.~5, pp. 2742--2759, 2022.

\bibitem{salzmann2020trajectron++}
T.~Salzmann, B.~Ivanovic, P.~Chakravarty, and M.~Pavone, ``Trajectron++: Dynamically-feasible trajectory forecasting with heterogeneous data,'' in \emph{Computer Vision--ECCV 2020: 16th European Conference, Glasgow, UK, August 23--28, 2020, Proceedings, Part XVIII 16}.\hskip 1em plus 0.5em minus 0.4em\relax Springer, 2020, pp. 683--700.

\bibitem{SEEM2023-pami}
D.~Wang, H.~Liu, N.~Wang, Y.~Wang, H.~Wang, and S.~McLoone, ``Seem: A sequence entropy energy-based model for pedestrian trajectory all-then-one prediction,'' \emph{IEEE Transactions on Pattern Analysis and Machine Intelligence}, vol.~45, no.~1, pp. 1070--1086, 2023.

\bibitem{gupta2018socialGAN}
A.~Gupta, J.~Johnson, L.~Fei-Fei, S.~Savarese, and A.~Alahi, ``Social gan: Socially acceptable trajectories with generative adversarial networks,'' in \emph{Proceedings of the IEEE Conference on Computer Vision and Pattern Recognition}, 2018, pp. 2255--2264.

\bibitem{li2021hierarchical}
Z.~Li, C.~Lu, Y.~Yi, and J.~Gong, ``A hierarchical framework for interactive behaviour prediction of heterogeneous traffic participants based on graph neural network,'' \emph{IEEE Transactions on Intelligent Transportation Systems}, vol.~23, no.~7, pp. 9102--9114, 2021.

\bibitem{li2024UQnet}
G.~Li, Z.~Li, V.~L. Knoop, and H.~{van Lint}, ``Unravelling uncertainty in trajectory prediction using a non-parametric approach,'' \emph{Transportation Research Part C: Emerging Technologies}, vol. 163, p. 104659, 2024.

\bibitem{gu2021densetnt}
J.~Gu, C.~Sun, and H.~Zhao, ``Densetnt: End-to-end trajectory prediction from dense goal sets,'' in \emph{Proceedings of the IEEE/CVF International Conference on Computer Vision}, 2021, pp. 15\,303--15\,312.

\bibitem{graph2022-its}
X.~Mo, Z.~Huang, Y.~Xing, and C.~Lv, ``Multi-agent trajectory prediction with heterogeneous edge-enhanced graph attention network,'' \emph{IEEE Transactions on Intelligent Transportation Systems}, vol.~23, no.~7, pp. 9554--9567, 2022.

\bibitem{vaswani2017Transformer}
A.~Vaswani, N.~Shazeer, N.~Parmar, J.~Uszkoreit, L.~Jones, A.~N. Gomez, {\L}.~Kaiser, and I.~Polosukhin, ``Attention is all you need,'' \emph{Advances in neural information processing systems}, vol.~30, 2017.

\bibitem{bae2024singulartrajectory-diff}
I.~Bae, Y.-J. Park, and H.-G. Jeon, ``Singulartrajectory: Universal trajectory predictor using diffusion model,'' in \emph{Proceedings of the IEEE/CVF Conference on Computer Vision and Pattern Recognition}, 2024, pp. 17\,890--17\,901.

\bibitem{2024MTR++_PAMI}
S.~Shi, L.~Jiang, D.~Dai, and B.~Schiele, ``Mtr++: Multi-agent motion prediction with symmetric scene modeling and guided intention querying,'' \emph{IEEE Transactions on Pattern Analysis and Machine Intelligence}, vol.~46, no.~5, pp. 3955--3971, 2024.

\bibitem{wang2017growing}
Y.-X. Wang, D.~Ramanan, and M.~Hebert, ``Growing a brain: Fine-tuning by increasing model capacity,'' in \emph{Proceedings of the IEEE Conference on Computer Vision and Pattern Recognition}, 2017, pp. 2471--2480.

\bibitem{mallya2018packnet}
A.~Mallya and S.~Lazebnik, ``Packnet: Adding multiple tasks to a single network by iterative pruning,'' in \emph{Proceedings of the IEEE conference on Computer Vision and Pattern Recognition}, 2018, pp. 7765--7773.

\bibitem{mallya2018piggyback}
A.~Mallya, D.~Davis, and S.~Lazebnik, ``Piggyback: Adapting a single network to multiple tasks by learning to mask weights,'' in \emph{Proceedings of the European conference on computer vision (ECCV)}, 2018, pp. 67--82.

\bibitem{kirkpatrick2017EWC}
J.~Kirkpatrick, R.~Pascanu, N.~Rabinowitz, J.~Veness, G.~Desjardins, A.~A. Rusu, K.~Milan, J.~Quan, T.~Ramalho, A.~Grabska-Barwinska \emph{et~al.}, ``Overcoming catastrophic forgetting in neural networks,'' \emph{Proceedings of the national academy of sciences}, vol. 114, no.~13, pp. 3521--3526, 2017.

\bibitem{li2017learningLwF}
Z.~Li and D.~Hoiem, ``Learning without forgetting,'' \emph{IEEE transactions on pattern analysis and machine intelligence}, vol.~40, no.~12, pp. 2935--2947, 2017.

\bibitem{lesort2019generative-dgr}
T.~Lesort, H.~Caselles-Dupr{\'e}, M.~Garcia-Ortiz, A.~Stoian, and D.~Filliat, ``Generative models from the perspective of continual learning,'' in \emph{2019 International Joint Conference on Neural Networks (IJCNN)}.\hskip 1em plus 0.5em minus 0.4em\relax IEEE, 2019, pp. 1--8.

\bibitem{lopez2017gradient-gem}
D.~Lopez-Paz and M.~Ranzato, ``Gradient episodic memory for continual learning,'' \emph{Advances in neural information processing systems}, vol.~30, 2017.

\bibitem{chaudhry2019efficient-agem}
A.~Chaudhry, R.~Marc'Aurelio, M.~Rohrbach, and M.~Elhoseiny, ``Efficient lifelong learning with a-gem,'' in \emph{7th International Conference on Learning Representations, ICLR 2019}.\hskip 1em plus 0.5em minus 0.4em\relax International Conference on Learning Representations, ICLR, 2019.

\bibitem{vitter1985random-rsvr}
J.~S. Vitter, ``Random sampling with a reservoir,'' \emph{ACM Transactions on Mathematical Software (TOMS)}, vol.~11, no.~1, pp. 37--57, 1985.

\bibitem{sae2014taxonomy}
S.~O.-R. A. V.~S. Committee \emph{et~al.}, ``Taxonomy and definitions for terms related to on-road motor vehicle automated driving systems,'' \emph{SAE Standard J}, vol. 3016, p.~1, 2014.

\bibitem{2025-humanguided-CL}
H.~Yang, Y.~Zhou, J.~Wu, H.~Liu, L.~Yang, and C.~Lv, ``Human-guided continual learning for personalized decision-making of autonomous driving,'' \emph{IEEE Transactions on Intelligent Transportation Systems}, pp. 1--0, 2025.

\bibitem{2022-EWC-ped-traj}
L.~Knoedler, C.~Salmi, H.~Zhu, B.~Brito, and J.~Alonso-Mora, ``Improving pedestrian prediction models with self-supervised continual learning,'' \emph{IEEE Robotics and Automation Letters}, vol.~7, no.~2, pp. 4781--4788, 2022.

\bibitem{GC2024-lifelong}
C.~Gong, C.~Lu, Z.~Li, Z.~Liu, J.~Gong, and X.~Chen, ``Beyond imitation: A life-long policy learning framework for path tracking control of autonomous driving,'' \emph{IEEE Transactions on Vehicular Technology}, vol.~73, no.~7, pp. 9786--9799, 2024.

\bibitem{shin2017continual-DGR}
H.~Shin, J.~K. Lee, J.~Kim, and J.~Kim, ``Continual learning with deep generative replay,'' \emph{Advances in neural information processing systems}, vol.~30, 2017.

\bibitem{li2023continual-survey}
Z.~Li, C.~Gong, Y.~Lin, G.~Li, X.~Wang, C.~Lu, M.~Wang, S.~Chen, and J.~Gong, ``Continual driver behaviour learning for connected vehicles and intelligent transportation systems: Framework, survey and challenges,'' \emph{Green Energy and Intelligent Transportation}, p. 100103, 2023.

\bibitem{lin2023rethinking}
Y.~Lin, Z.~Li, C.~Gong, Q.~Liu, C.~Lu, and J.~Gong, ``Rethinking trajectory prediction in real-world applications: An online task-free continual learning perspective,'' in \emph{2023 IEEE 26th International Conference on Intelligent Transportation Systems (ITSC)}.\hskip 1em plus 0.5em minus 0.4em\relax IEEE, 2023, pp. 5020--5026.

\bibitem{he2024dyson}
Y.~He, Y.~Chen, Y.~Jin, S.~Dong, X.~Wei, and Y.~Gong, ``Dyson: Dynamic feature space self-organization for online task-free class incremental learning,'' in \emph{Proceedings of the IEEE/CVF Conference on Computer Vision and Pattern Recognition}, 2024, pp. 23\,741--23\,751.

\bibitem{aljundi2019-gss}
R.~Aljundi, M.~Lin, B.~Goujaud, and Y.~Bengio, ``Gradient based sample selection for online continual learning,'' \emph{Advances in neural information processing systems}, vol.~32, 2019.

\bibitem{2015Distilling}
G.~Hinton, O.~Vinyals, and J.~Dean, ``Distilling the knowledge in a neural network,'' \emph{Computer Science}, vol.~14, no.~7, pp. 38--39, 2015.

\bibitem{zhan2019interaction-dataset}
W.~Zhan, L.~Sun, D.~Wang, H.~Shi, A.~Clausse, M.~Naumann, J.~Kummerle, H.~Konigshof, C.~Stiller, A.~de~La~Fortelle \emph{et~al.}, ``Interaction dataset: An international, adversarial and cooperative motion dataset in interactive driving scenarios with semantic maps,'' \emph{arXiv preprint arXiv:1910.03088}, 2019.

\bibitem{2020focal_loss}
T.-Y. Lin, P.~Goyal, R.~Girshick, K.~He, and P.~Dollár, ``Focal loss for dense object detection,'' \emph{IEEE Transactions on Pattern Analysis and Machine Intelligence}, vol.~42, no.~2, pp. 318--327, 2020.

\bibitem{gilles2022gohome}
T.~Gilles, S.~Sabatini, D.~Tsishkou, B.~Stanciulescu, and F.~Moutarde, ``Gohome: Graph-oriented heatmap output for future motion estimation,'' in \emph{2022 International Conference on Robotics and Automation (ICRA)}.\hskip 1em plus 0.5em minus 0.4em\relax IEEE, 2022, pp. 9107--9114.

\bibitem{kingma2014adam}
D.~P. Kingma and J.~Ba, ``Adam: A method for stochastic optimization,'' \emph{arXiv preprint arXiv:1412.6980}, 2014.

\bibitem{li2021privacy}
Y.~Li, X.~Tao, X.~Zhang, J.~Liu, and J.~Xu, ``Privacy-preserved federated learning for autonomous driving,'' \emph{IEEE Transactions on Intelligent Transportation Systems}, vol.~23, no.~7, pp. 8423--8434, 2021.

\end{thebibliography}

\vspace{-10 mm}
\begin{IEEEbiography}[{\includegraphics[width=1in,height=1.25in,clip,keepaspectratio]{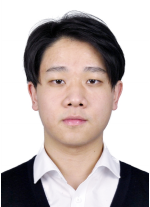}}]{Yunlong Lin}
received the B.S. degree in mechanical engineering from Beijing Institute of Technology (BIT), Beijing, China, in 2022. He is currently pursuing the Ph.D. degree in BIT. His research focuses on interactive behavior modeling, trajectory prediction, continual learning, and decision-making of intelligent vehicles.
\end{IEEEbiography}
\vspace{-10 mm}
\begin{IEEEbiography}[{\includegraphics[width=1in,height=1.25in,clip,keepaspectratio]{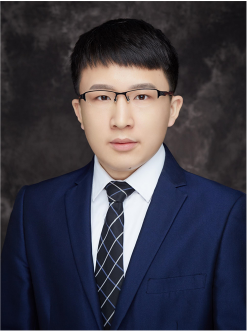}}]{Zirui Li}
received the B.S. and  Ph.D. degree in mechanical engineering from Beijing Institute of Technology (BIT), Beijing, China in 2019 and 2025, respectively. He is currently a postdoctoral research fellow in Nanyang Technological University. From June, 2021 to July, 2022, he was a visiting researcher in Delft University of Technology (TU Delft). From Aug, 2022 to Jun, 2024. He was the visiting researcher in the Chair of Traffic Process Automation at the Faculty of Transportation and Traffic Sciences “Friedrich List” of the TU Dresden. His research focuses on interactive behavior modeling, risk assessment and motion planning of automated vehicles.
\end{IEEEbiography}
\vspace{-10 mm}
\begin{IEEEbiography}[{\includegraphics[width=1in,height=1.25in,clip,keepaspectratio]{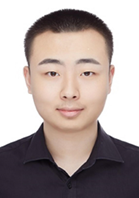}}]{Guodong Du}
received the B.S. degree in Mechanical Engineering from Beijing Institute of Technology, Beijing, China, in 2019, and the Ph.D. degree in Automobile Engineering from Beijing Institute of Technology, Beijing, China, in 2025.  He also serves as an academic guest of ETH Zurich. His research interests include motion planning and control, reinforcement learning algorithm, energy management of hybrid electric vehicles.s
\end{IEEEbiography}
\vspace{-10 mm}
\begin{IEEEbiography}[{\includegraphics[width=1in,height=1.25in,clip,keepaspectratio]{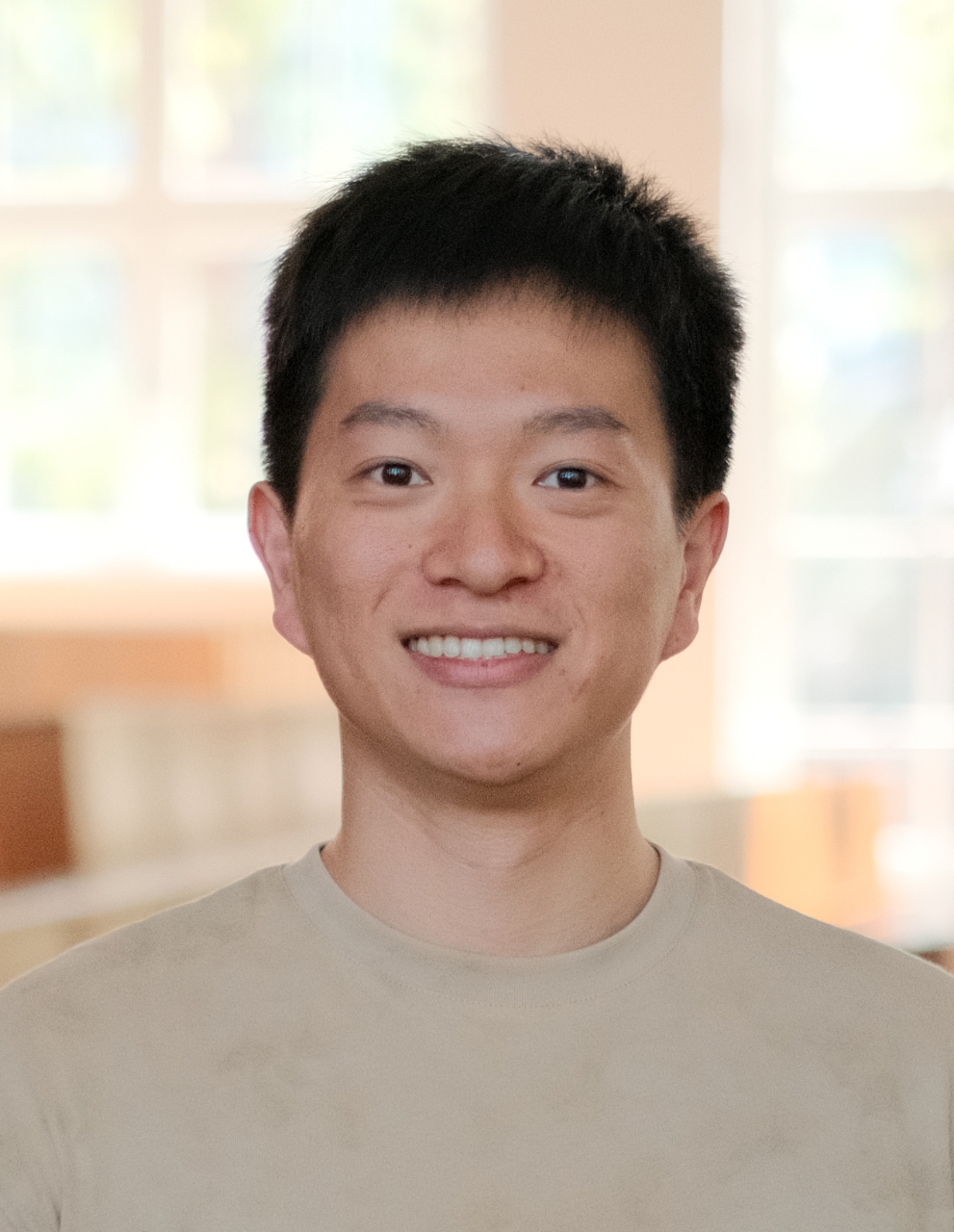}}]{Xiaocong Zhao}
 received the Ph.D. degree in Transportation Engineering from Tongji University, Shanghai, China. He is currently a postdoctoral research fellow in the College of Transportation at Tongji University, Shanghai, China. From Nov 2022 to Oct 2023, he was a visiting researcher with the Chair of Traffic Process Automation, Faculty of Transportation and Traffic Sciences “Friedrich List”of  Technische Universität Dresden, Germany. His main research interests include human-machine interaction, social driving behaviors modeling, and interactive decision-making.
\end{IEEEbiography}
\vspace{-10 mm}
\begin{IEEEbiography}[{\includegraphics[width=1in,height=1.25in,clip,keepaspectratio]{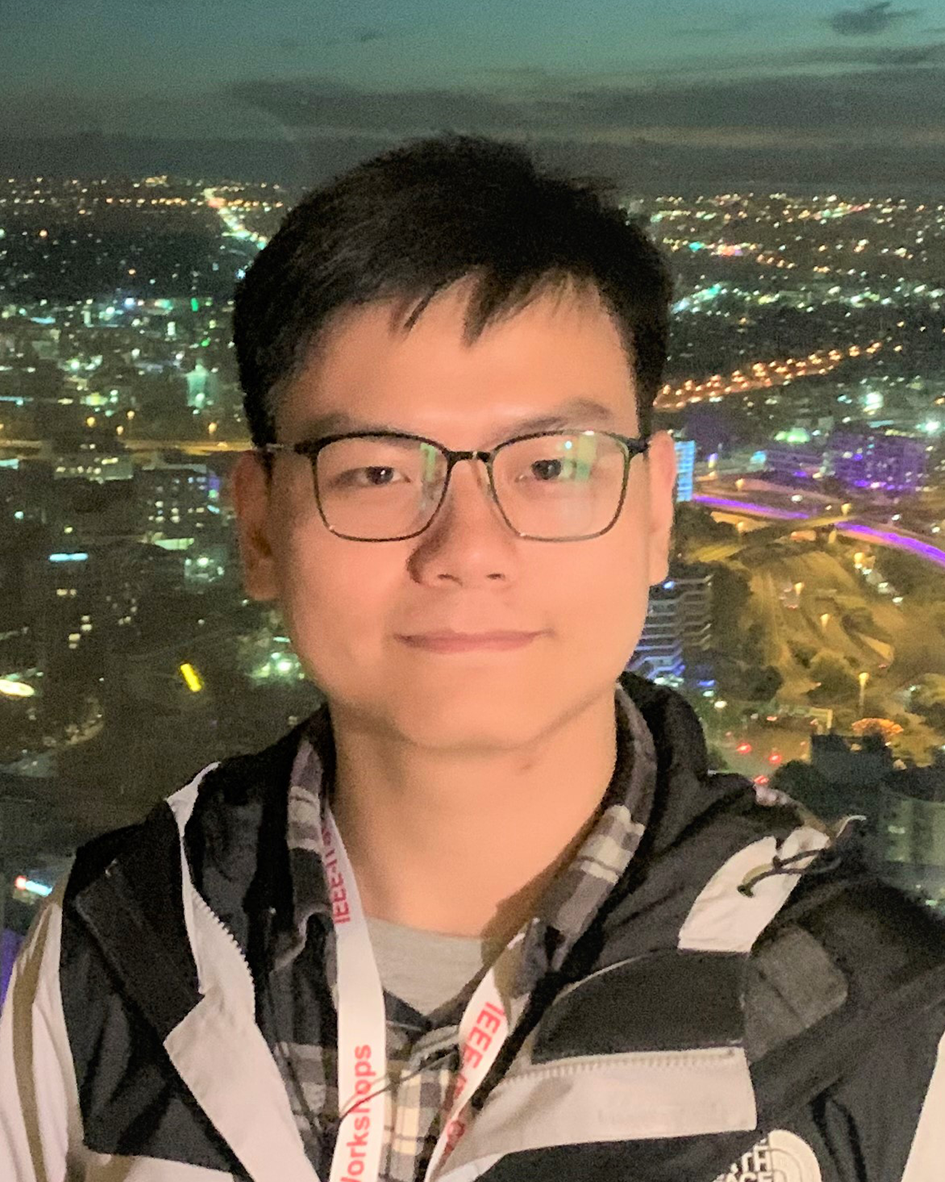}}]{Cheng Gong}
received the B.S. degree in mechanical engineering from Beijing Institute of Technology (BIT), China, in 2020. He is currently pursuing the Ph.D. degree in BIT. His research interests include intelligent vehicles, motion planning and control, and lifelong machine learning.
\end{IEEEbiography}
\vspace{-10 mm}
\begin{IEEEbiography}[{\includegraphics[width=1in,height=1.25in,clip,keepaspectratio]{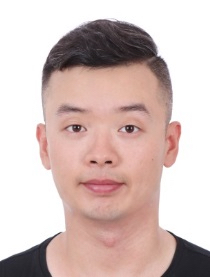}}]{Xinwei Wang}
is a Lecturer (Assistant Professor) at Queen Mary University of London (QMUL), UK. He was a Postdoc at TU Delft, The Netherlands from 2020 to 2022 and at QMUL from 2019 to 2020, respectively. He obtained a PhD degree from Beihang University, China in 2019. Over the years, he has integrated computational intelligence, machine learning and systems engineering for risk assessment, motion planning and decision making in intelligent systems. He is a recipient of the Marie Sklodowska-Curie Actions Co-Fund Fellowship (2022), and IEEE ITSS Young Professionals Travelling Fellowship (2022).
\end{IEEEbiography}
\vspace{-10 mm}
\begin{IEEEbiography}[{\includegraphics[width=1in,height=1.25in,clip,keepaspectratio]{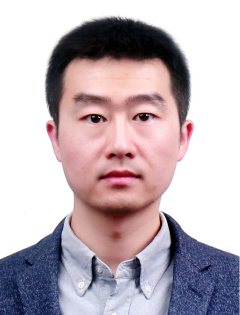}}]{Chao Lu}
received the B.S. degree in transport engineering from the Beijing Institute of Technology (BIT), Beijing, China, in 2009 and the Ph.D. degree in transport studies from the University of Leeds, Leeds, U.K., in 2015. In 2017, he was a Visiting Researcher with the Advanced Vehicle Engineering Centre, Cranfield University, Cranfield, U.K. He is currently an Associate Professor with the School of Mechanical Engineering, BIT. His research interests include intelligent transportation and vehicular systems, driver behavior modeling, reinforcement learning, and transfer learning and its applications.
\end{IEEEbiography}
\vspace{-10 mm}
\begin{IEEEbiography}[{\includegraphics[width=1in,height=1.25in,clip,keepaspectratio]{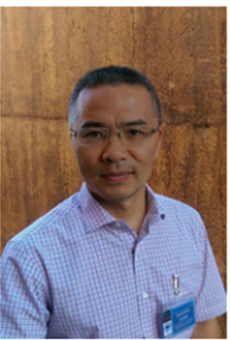}}]{Jianwei Gong}
received the B.S. degree from the National University of Défense Technology, Changsha, China, in 1992, and the Ph.D. degree from Beijing Institute of Technology (BIT), Beijing, China, in 2002. Between 2011 and 2012, he was a Visiting Scientist with the Robotic Mobility Group, Massachusetts Institute of Technology, Cambridge, MA, USA. He is currently a Professor with the School of Mechanical Engineering, BIT. His research interests include intelligent vehicle environment perception and understanding, decision making, path/motion planning, and control.
\end{IEEEbiography}

\end{document}